\documentclass[lettersize,journal]{IEEEtran}

\usepackage{algorithmic}
\usepackage{algorithm}
\usepackage{array}
\usepackage{textcomp}
\usepackage{stfloats}
\usepackage{url}
\usepackage{verbatim}
\usepackage{graphicx}
\usepackage{xcolor}
\usepackage{amsmath}
\usepackage{amsfonts}
\usepackage{booktabs}
\usepackage{adjustbox}
\usepackage{lipsum}
\usepackage{balance}
\usepackage[doi=false,isbn=false,url=true,date=year,backend=biber,style=ieee,sorting=none,maxnames=3,minnames=1]{biblatex} 
\AtBeginBibliography{\small}
\usepackage{orcidlink}
\usepackage{subcaption}
\usepackage{enumitem}

\newcommand{\node}{\mathbf{n}}
\newcommand{\expected}{\hat{\mathbf{z}}_{ij}}

\newcommand{\Lim}[1]{\raisebox{0.5ex}{\scalebox{0.8}{$\displaystyle \lim_{#1}\;$}}}

\DeclareSourcemap{ 
    \maps[datatype=bibtex]{
      \map{
      \step[fieldsource = booktitle,
          match = Conference on Computer Vision and Pattern Recognition,
          replace = CVPR]
        \step[fieldsource = booktitle,
          match = Computer Vision and Pattern Recognition,
          replace = CVPR]          
        \step[fieldsource = booktitle,
          match = International Conference on Computer Vision,
          replace = ICCV]
        \step[fieldsource = journal,
          match = Intelligent Transportation Systems,
          replace = ITS]
        \step[fieldsource = booktitle,
          match = Intelligent Vehicles Symposium,
          replace = IV Symp.]
        \step[fieldsource = journal,
          match = Intelligent Vehicles Symposium,
          replace = IV Symp.]
        \step[fieldsource = booktitle,
          match = International Conference on Intelligent Transportation Systems,
          replace = ITSC]
        \step[fieldsource = journal,
          match = International Conference on Intelligent Transportation Systems,
          replace = ITSC]
        \step[fieldsource = journal,
              match = Transactions,
              replace = Trans.]
        \step[fieldsource = booktitle,
              match = {International Conference on Automation, Robotics and Applications},
              replace = ICARA]
      \step[fieldsource = journal,
              match = {Systems, Man, and Cybernetics},
              replace = SMC]
        \step[fieldsource = journal,
              match = {Systems, Man and Cybernetics},
              replace = SMC]
        \step[fieldsource = booktitle,
              match = {International Conference on Multisensor Fusion and Integration for Intelligent Systems},
              replace = MFI]
        \step[fieldsource = booktitle,
              match = {International Conference on Intelligent Robots and Systems},
              replace = IROS]
        \step[fieldsource = journal,
              match = {Robotics and Automation Letters},
              replace = RA-L]
        \step[fieldsource = booktitle,
        match = Conference on robot learning,
        replace = CoRL
        ]
        \step[fieldsource = booktitle,
        match = {International Convention on Information, Communication and Electronic Technology},
        replace = MIPRO
        ]
        \step[fieldset=publisher, null]
        }
    }      
}

\AtEveryBibitem{
    \ifentrytype{misc}{\iflistundef{publisher}{}{\clearfield{url}}}{\clearfield{url}}
}

\newcommand{\subfigWidth}{0.3\textwidth}

\begin{document}

\title{SDS++: Online Situation-Aware Drivable Space Estimation for Automated Driving}
\author{Manuel~Muñoz~Sánchez\orcidlink{0000-0001-5997-9039},
        Gijs Trots,
        Robin Smit\orcidlink{0009-0001-8039-5457},
        Pedro Vieira Oliveira,
        Emilia~Silvas\orcidlink{0000-0002-3038-7053},\\
        Jos~Elfring\orcidlink{0000-0001-7277-9904},
        and~René~van~de~Molengraft\orcidlink{0000-0002-5095-4297
}
\thanks{
This work was supported by the SAFE-UP project under EU's Horizon 2020 research and innovation programme, grant agreement 861570 and the DITM project funded by NextGenerationEU, Ministerie van Infrastructuur en Waterstaat, RvO, grant agreement NGFDI2201. 
\textit{(Corresponding author: Manuel Muñoz Sánchez.)}
}%
\thanks{M. Muñoz Sánchez, G. W. Trots, E. Silvas, J. Elfring and R. van de Molengraft are with the Department of Mechanical Engineering, Eindhoven University of Technology, Eindhoven, The Netherlands. (e-mail: m.munoz.sanchez@tue.nl).}
\thanks{R. Smit and P. Vieira Oliveira are with the Department of Integrated Vehicle Safety, TNO, Helmond, The Netherlands.}
\thanks{E. Silvas is also with the Department of Integrated Vehicle Safety, TNO, Helmond, The Netherlands.}
}

\markboth{Submitted to IEEE Transactions on Intelligent Vehicles}{}


\maketitle

\begin{abstract}
Autonomous Vehicles (AVs) need an accurate and up-to-date representation of the environment for safe navigation. Traditional methods, which often rely on detailed environmental representations constructed offline, struggle in dynamically changing environments or when dealing with outdated maps. Consequently, there is a pressing need for real-time solutions that can integrate diverse data sources and adapt to the current situation. An existing framework that addresses these challenges is SDS (situation-aware drivable space). However, SDS faces several limitations, including its use of a non-standard output representation, its choice of encoding objects as points, restricting representation of more complex geometries like road lanes, and the fact that its methodology has been validated only with simulated or heavily post-processed data. This work builds upon SDS and introduces SDS++, designed to overcome SDS's shortcomings while preserving its benefits. SDS++ has been rigorously validated not only in simulations but also with unrefined vehicle data, and it is integrated with a model predictive control (MPC)-based planner to verify its advantages for the planning task. The results demonstrate that SDS++ significantly enhances trajectory planning capabilities, providing increased robustness against localization noise, and enabling the planning of trajectories that adapt to the current driving context. 

\end{abstract}


\section{Introduction}

Automated driving (AD) promises to enhance road safety and efficiency~\cite{Yurtsever2020ATechnologies, Milakis2017PolicyResearch}. To fulfill this promise, an automated vehicle (AV) must accurately determine the precise space on which it is allowed to drive, i.e. the \textit{drivable space}~\cite{MunozSanchez2022}. In AD applications, it is common to rely on high-definition (HD) maps to determine the drivable space and navigate the environment~\cite{Liu2020HighAnalysis,Khoche2022SemanticDriving,EbrahimiSoorchaei2022High-DefinitionVehicles,Charroud2024LocalizationSurvey,Jo2018SimultaneousCar}. However, HD maps, typically generated offline~\cite{Liu2020HighAnalysis, Charroud2024LocalizationSurvey}, must be updated online to reflect current environments and support situation-aware decision making in challenging driving scenarios~\cite{Yurtsever2020ATechnologies,MunozSanchez2022,Gerwien2021TowardsDriving}.

%
Fig.~\ref{fig:examples} shows four examples of such challenging scenarios for an AV. 
In Fig.\ref{fig:examples-boulder}, a boulder blocks the lane\footnote{From online news article: \url{https://bit.ly/boulder-blocking-lane}}, requiring a rule violation by crossing a solid yellow line to continue. 
Fig.\ref{fig:examples-roundabout} shows a new roundabout not yet updated in the map\footnote{In Google Maps: \url{https://bit.ly/maps-street-view-missing-roundabout}}. 
Fig.~\ref{fig:examples-markings} shows an unusual road marking design\footnote{From online news article: \url{https://bit.ly/challenging-road-markings}} that would be challenging for an AV. Lastly, 
Fig.\ref{fig:examples-parking} depicts a parking lot with implicit driving and parking rules despite the absence of lane markings\footnote{Online: \url{https://bit.ly/gravel-parking-lot}}. 
\renewcommand{\subfigWidth}{0.4935\linewidth}
\begin{figure}[tb]
    \centering
    \begin{subfigure}[t]{\subfigWidth}
      \centering
        \includegraphics[width=\textwidth]{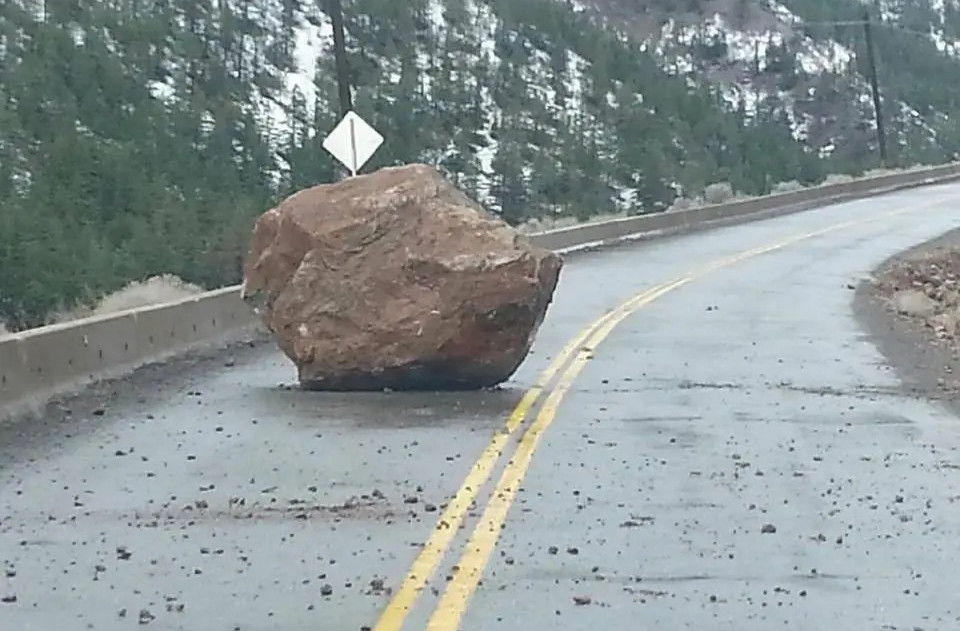}
        \caption{Lane blocked by boulder.}\label{fig:examples-boulder}
    \end{subfigure}
    \begin{subfigure}[t]{\subfigWidth}
         \centering
        \includegraphics[width=\textwidth]{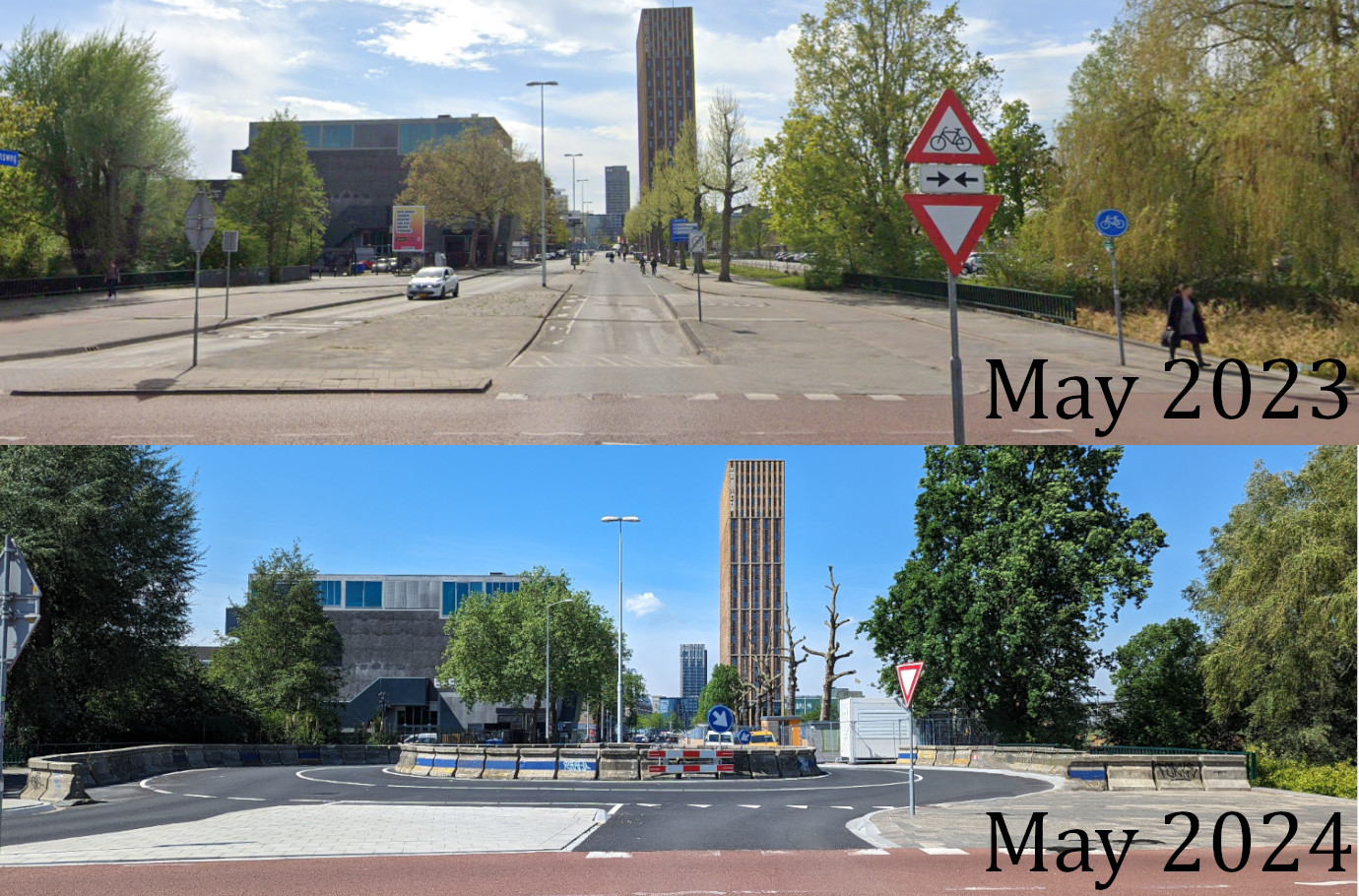}
        \caption{New roundabout not present in outdated map.}\label{fig:examples-roundabout}
    \end{subfigure}  
    \begin{subfigure}[t]{\subfigWidth}
      \centering
        \includegraphics[width=\textwidth]{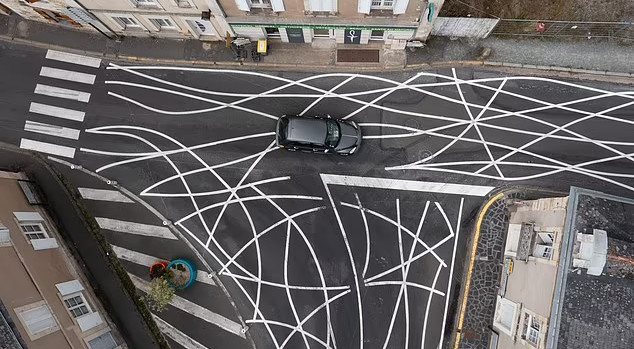}
        \caption{Challenging road markings.}\label{fig:examples-markings}
    \end{subfigure}
    \begin{subfigure}[t]{\subfigWidth}
      \centering
        \includegraphics[width=\textwidth]{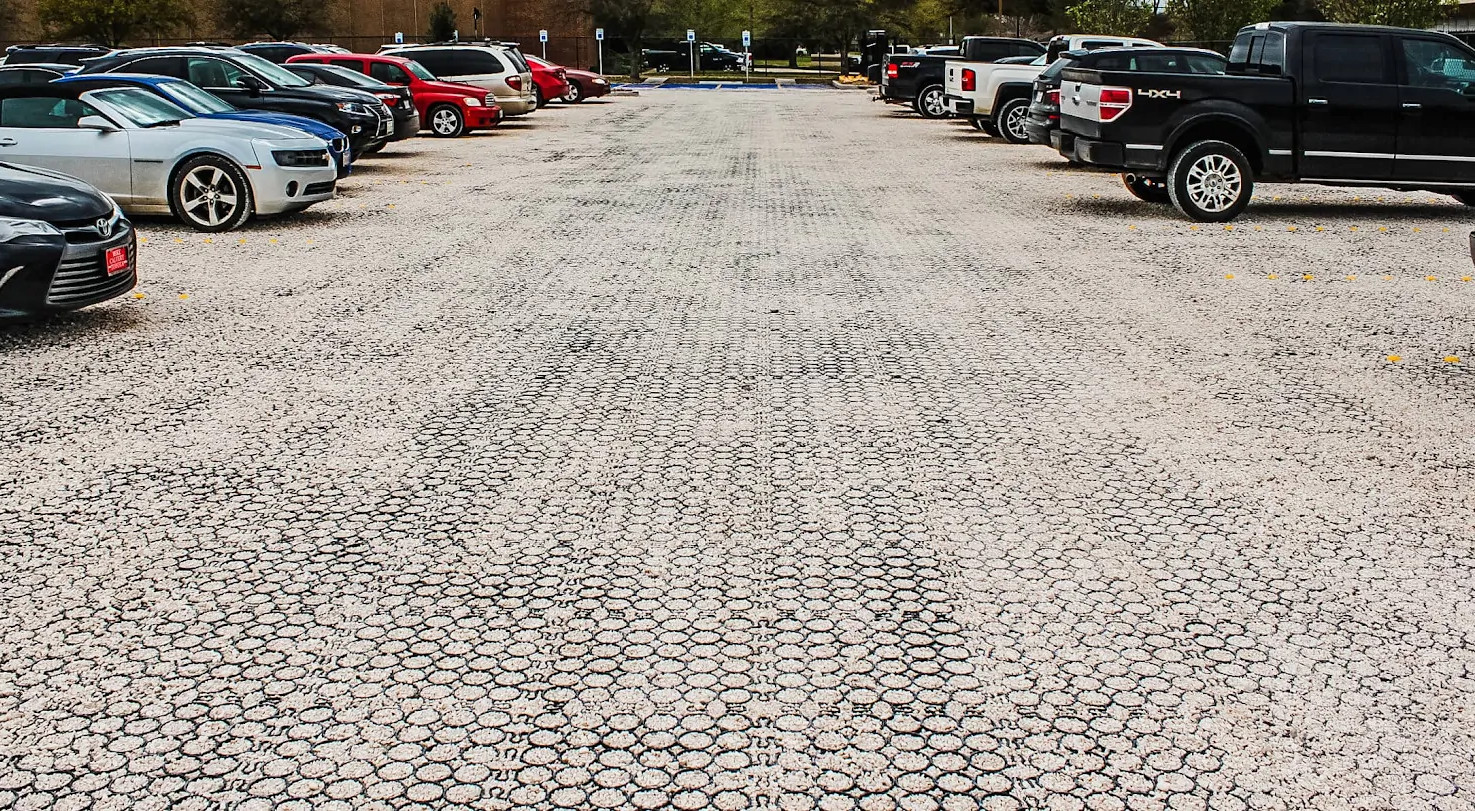}
        \caption{Parking lot without markings.}\label{fig:examples-parking}
    \end{subfigure}
    \caption{ Examples of challenging driving situations for an AV. }
    \label{fig:examples}
\end{figure}
Without the ability to adapt to current scenarios, AVs might struggle to maintain safe operation in such conditions. For instance, although typically not allowed, crossing a solid lane line might be necessary when an obstacle blocks the lane (Fig.~\ref{fig:examples-boulder}).

Maintaining up-to-date, computationally intensive HD maps poses challenges in dynamic environments and when dealing with outdated maps, requiring real-time, in-vehicle solutions that integrate diverse data sources and adapt to current driving situations~\cite{Charroud2024LocalizationSurvey, Yurtsever2020ATechnologies}.
A methodology addressing these challenges should allow efficient updates to maintain a correct drivable space representation, be sufficiently flexible to support a wide variety of AV architectures, and allow reasoning about the drivability of the space to enable situation-aware decision making. Thus, an effective drivable space estimation methodology should satisfy the following criteria:

\paragraph{Storage \& computation efficiency}
A representation that efficiently captures only essential information reduces storage needs and computational power requirements, facilitating real-time execution

\paragraph{Sensor agnosticism} 
Environment modeling methods often rely on specific rich sensor data (e.g. LiDAR or cameras). This dependence limits the method to AVs with the same sensor types and requires re-calibration if sensors change. Avoiding reliance on specific sensors additionally enhances robustness against individual sensor failures.

\paragraph{Drivability reasoning \& adaptability}
In the AD domain there is a multitude of assumptions that can be made regarding the drivability of an area, i.e. domain knowledge assumptions. For instance, if a vehicle is seen driving but the surface beneath is not visible, it is assumed to be drivable. Similarly, crossing a solid line to avoid an obstacle can be acceptable in some situations (Fig.~\ref{fig:examples-boulder}). A method allowing incorporation of these assumptions, which are intuitive for human drivers, enhances and adapts drivable space estimates.


\paragraph{Compatibility with existing planning algorithms} 

The provided environment representation  should be compatible with the representations typically used by state-of-the-art planners. A standardized format eases integration and fully leverages situation-aware drivable space estimation methodologies.

The SDS (situation-aware drivable space) framework~\cite{MunozSanchez2022} estimates drivable space from detected objects and their semantic information, using a graph-based SLAM optimization approach~\cite{Grisetti2010ASLAM} to create a probabilistic graph representation of the most likely drivable space configuration. SDS meets most of the criteria for effective drivable space estimation, while offering two key benefits: it incorporates more information sources than traditional methods and interfaces only with late fusion-detected objects, enhancing robustness against sensor failures and allowing seamless sensor setup modifications. It also integrates domain knowledge assumptions based on semantic information, allowing for drivable space estimation even when geometric details are missing from the sources. Nevertheless, SDS has limitations that hinder its broader applicability:

\setcounter{paragraph}{0}

\paragraph{Non-standardized output representation} 
SDS opted for a graph-based representation which, while informative, presents challenges for integration into AV planners due to the lack of a standardized format.

\paragraph{Limited to basic shapes} 
SDS encodes objects as point features, complicating the preservation of complex geometries and curves, such as road lanes.

\paragraph{Validation on simulated and post-processed data} 
SDS has been predominantly validated in simulations and heavily post-processed datasets, which may not reflect the accuracy and cleanliness typically achievable by AVs in real-world scenarios.

This paper introduces SDS++, which builds upon SDS and overcomes its limitations while retaining its main advantages. SDS++ transitions from graph-based representations to Artificial Potential Fields (APFs), offering a lightweight yet expressive means to describe the drivable space. This shift enables a more standardized output readily compatible with state-of-the-art AV planning systems~\cite{vanderPloeg2022LongApproach}.

SDS++ represents objects with APFs using a combination of implicit and sigmoid functions~\cite{Ren2007AFunctions}, providing a more accurate representation than the commonly used bi-Gaussian fields~\cite{vanderPloeg2022LongApproach, Hongyu2018AnDriving, Jia2022Car-followingField, Huang2020AApproach}. To address the limitation of handling only point features, a custom implicit function factor is introduced for graph-based SLAM optimization. This enhancement allows the incorporation of complex geometries, such as partially detected lines, enriching the environmental model and enhancing accuracy. In AD environments with sparse discriminative features, lines capture the boundary between different surfaces, enhancing accurate environment representation and robustness. 



SDS++ maintains the benefits of integrating domain knowledge assumptions, resulting in a flexible drivable space estimation methodology that adapts to current situations. When integrated with motion planning, the AV can navigate challenging scenarios that traditional environment representations do not support. Our contributions include:
\begin{itemize}
    \item An improved drivable space estimation method that addresses SDS's main limitations while maintaining its lightweight representation and domain knowledge support to adapt to current driving situations.
    
    \item The introduction of a factor for direct optimization of (partially observed) line features described by implicit functions in graph-based SLAM, supporting shapes described by multivalued functions and enhancing accurate environment representation and robustness.    
    
    \item Drivability extraction from detected objects and their semantic information through APFs, encoding the geometry of environmental elements more accurately than the commonly used bi-Gaussian representation.
    
    \item Validation of the proposed SDS++ with real-world driving data collected with an AD-enabled vehicle.
\end{itemize}

The remainder of this article is structured as follows. Section~\ref{sec:background} introduces concepts upon which our proposed method is built. Section~\ref{sec:methodology} introduces SDS++, our improved drivable space estimation method. Section~\ref{sec:experiments} outlines the experiments performed to validate SDS++ and their results. Finally, Section~\ref{sec:conclusion} concludes the work and outlines future improvements.

\section{Background}\label{sec:background}
\begin{figure*}[b]
    \centering
    \includegraphics[width=0.93\linewidth]{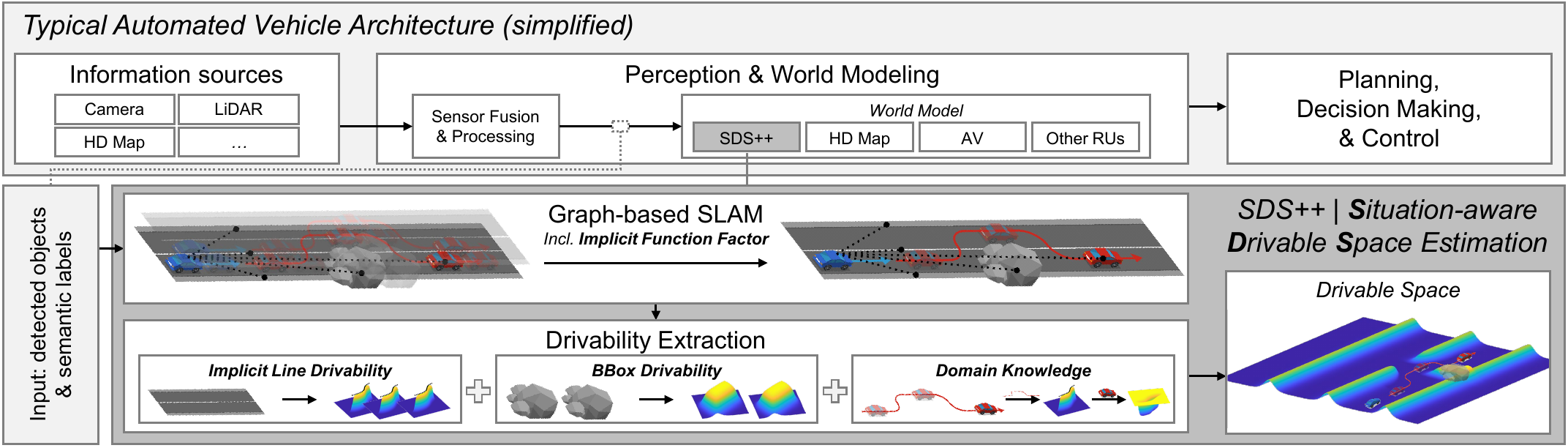}
    \caption{Typical architecture of an AV (simplified), featuring our proposed online situation-aware drivable space (SDS++) estimation. Raw input data is processed to build a world model, which is then used for planning and executing the best course of action. Our SDS++ enables the construction of a situation-aware drivable space and ego localization within it, allowing the AV to navigate complex scenarios that traditional environment representations cannot handle.}
    \label{fig:overview}
\end{figure*}

\subsection{Graph-based SLAM}\label{subsec:background-slam}
Graph-based simultaneous localization and mapping (SLAM) is frequently used to localize a mobile robot in a partially or fully unknown environment~\cite{Thrun2002ProbabilisticRobotics,Grisetti2010ASLAM}, and it is commonly used in AD applications to build or update a map online~\cite{Charroud2024LocalizationSurvey,Yurtsever2020ATechnologies,Jo2018SimultaneousCar}. Graph-based SLAM methods consist of two main components: a front-end, and a back-end~\cite{Cadena2016PastAge}. The front-end first processes sensor data to create a graph $\mathbf{G}=(\mathbf{N}, \mathbf{E})$ consisting of $|\mathbf{N}|$ nodes and $|\mathbf{E}|$ edges, where a node $\node \in \mathbf{N}$ encodes robot poses or landmark positions, and an edge $\mathbf{e}_{ij} \in \mathbf{E} $ between nodes
$\node_{i}$ and $\node_{j}$ encodes an expected spatial constraint, $\expected$, between the nodes. The actual measurements, $\mathbf{z}_{ij}$, can originate from sensor measurements. Following the construction of the optimization graph, the back-end optimizer determines $\mathbf{N^*}$, the configuration of nodes that best aligns with the relative spatial constraints in the edges. This optimization aims to minimize $\mathbf{\varepsilon}\left(\node_{i}, \node_{j}\right)=\mathbf{z}_{ij}-\expected$, the error  between the expected and actual measurements of nodes $\mathbf{n}_i$ and $\mathbf{n}_j$, denoted as $\mathbf{\varepsilon}_{ij}$ for simplicity of notation. That is, 
\begin{equation}\label{eq:slam-error-minimization}
    \mathbf{N}^{*}=\underset{\mathbf{N}}{\operatorname{argmin}} \sum_{\mathbf{e}_{ij} \in \mathbf{E}} \mathbf{\varepsilon}_{i j}^{T} \mathbf{\Omega}_{i j} \mathbf{\varepsilon}_{ij},
\end{equation} 
where $\mathbf{\Omega}_{i j}$ denotes the information matrix representing the certainty of the constraint. This matrix is derived from the measurement variance and is equal to the inverse of the covariance matrix

Traditional graph-based SLAM methods predominantly focus on point features. However, advancements in the field have led to the inclusion of additional structures commonly observed in man-made environments, such as straight lines~\cite{Ruifang2017Line-basedSLAM,Zhou2021ImprovedEnvironments} and planes~\cite{Arndt2020FromGraphs,Aloise2019SystematicOptimization}.
These elements, however, do not accurately capture the geometry of other non-straight structures also commonly found in AD environments, such as curved road lanes. To address this limitation, SDS++ incorporates a novel custom factor specifically designed to optimize possibly curved line features.

\subsection{Artificial Potential Fields for Environment Representation}

There is a wide variety of representations commonly used in AD applications, such as HD maps, feature maps, point clouds, grid maps, and APFs~\cite{EbrahimiSoorchaei2022High-DefinitionVehicles,Khoche2022SemanticDriving,Hongyu2018AnDriving}. These representations offer varying degrees of detail and computational demands, each suited to different applications.

Among these, APFs stand out for their unique approach to navigation, treating it as moving through a field of forces where obstacles generate repulsive forces and goals create attractive forces. This method enables efficient path planning and obstacle avoidance. However, APFs are not the most popular choice to model the environment due to their limited ability to capture detailed environmental semantic information, and the possibility to get stuck in local minima when using them for path planning. Despite these limitations, they have been applied successfully for navigation in a multitude of scenarios including car following~\cite{Jia2022Car-followingField}, overtaking~\cite{Lu2020HybridDriving}, crossing pedestrian avoidance~\cite{Bhatt2023MPC-PF:Fields,munozsanchez2024prediction}, and risk estimation of occluded areas~\cite{vanderploeg2023overcoming}. APFs typically model environmental elements as bi-Gaussian fields, which do not capture their shape accurately~\cite{Ji2023TriPField:Vehicles}, complicating their use in safety-critical applications such as AD.


Building on the foundation laid by SDS, our aim is to develop a minimal environmental representation that captures essential information for safe navigation. This representation will not replace the detailed mapping methodology of the AV, but aid in the detection and correction of inconsistencies with the currently known map~\cite[Fig.~1]{MunozSanchez2022}, and enable continued safe operation of the vehicle.

APFs are selected for their lightweight and continuous nature.
Additionally, APFs allow straightforward integration of domain knowledge by superposition of additional APFs, and their close link to navigation facilitates integration with already existing planning methods.



\section{Methodology}\label{sec:methodology}


This section details the improved methodology for online SDS estimation, which we named SDS++. Fig.~\ref{fig:overview} shows a high-level diagram of the data flow in SDS++ and where it stands within a typical modular AV architecture. 
To avoid reliance on any specific information source, SDS++ interfaces directly with fused objects and their semantic information. Since SDS++ does not make strong assumptions about the quality of the information sources or fusion algorithms, it initially obtains the most likely configuration of the environment using all available information to formulate a graph-based SLAM optimization, incorporating a novel implicit function factor as detailed in Section~\ref{subsec:slam}. Next, APFs representing the drivability of the space are extracted from each object, and in combination with domain knowledge assumptions SDS++ provides a drivable space that is consistent with the current driving situation, as explained in Section~\ref{subsec:drivability}.

\subsection{Graph-based SLAM \& Implicit Function Factor} \label{subsec:slam}
To ensure an up-to-date representation of the environment is available, a graph-based SLAM optimization is formulated to construct a local representation of the space around the AV.\footnote{Implemented with GTSAM~\cite{Dellaert2017FactorPerception,Dellaert2022Borglab/gtsam}}. Essentially, the formulation is a standardized landmark-based SLAM problem~\cite{Grisetti2010ASLAM}, which optimizes the AV's trajectory while simultaneously optimizing over the observed landmarks, as introduced in Section~\ref{subsec:background-slam}. In addition to the standard formulation, which mainly deals with robot poses and point landmarks, we introduce a custom factor to optimize over partially observed lines, as detailed in Section~\ref{subsec:implicit-function-lines}.

\subsubsection{Building \& Optimizing the Graph}\label{{subsec:building-graph}}
A node $\node \in \mathbf{N}$ in the graph encodes a detected object and has additional attributes,
\begin{equation} \label{eq:drivable-space-node}
  \node = \left[\tau, \mathbf{x}, \mathbf{y}, \theta_{opt}, \mathcal{S}_{opt}\right],
\end{equation}
where  $\tau \in \{AV, LM, LL \}$  denotes whether this node originates from an object to be encoded as a point feature, such as the pose of the vehicle (AV) and an observed landmark (LM), or from an object to be encoded as a line feature, such as an observed lane\footnote{Note that, although we only encode lane lines, any lines can be encoded in the same manner.} (LL).  If $\tau \in \{AV, LM\}$, $\mathbf{x}$ and $\mathbf{y}$ denote the single position of the encoded feature, whereas if $\tau = LL$, $\mathbf{x}$ and $\mathbf{y}$ denote a sequence of points from the line feature\footnote{We assume lane measurements are available as a collection of sampled points from the line representing the lane, or in a shape where sampled points can be extracted.}. Additional parameters such as the orientation $\theta$ and other semantic information $\mathcal{S}$ are optional, as denoted by $opt$, since this information is not always available or applicable to every node. 
To simplify notation, we refer to the attribute of a node with superscripts, and subscripts are used to refer to a specific node. For instance, to refer to an attribute $\psi \in \{\tau, \mathbf{x}, \mathbf{y}, \theta, \mathcal{S}\}$ of the $i^{th}$ node, it is denoted as $\node^\psi_i$. Similarly, we denote the nodes and edges of graph $\mathbf{G}$ as $\mathbf{G}^\mathbf{N}$ and $\mathbf{G}^\mathbf{E}$ respectively.


Edges encoding expected spatial constraints are introduced to the graph as previously described in Section~\ref{subsec:background-slam}. Depending on the type of nodes an edge connects, the constraint and resulting error can originate from different sources. 
For instance AV-AV constraints can be extracted from odometry measurements, AV-LM constraints from landmark measurements, and AV-LL from lane measurements. 




Finally, the optimization defined by~\eqref{eq:slam-error-minimization} is applied to obtain the optimal configuration of the nodes encoding all detected objects\footnote{Note that a time-based sliding window is applied to discard nodes arising from old observations, and only the most recent ones that fall within a predefined time window are used for the graph optimization.}. 
This process is standard for point features~\cite{Grisetti2010ASLAM}, and to optimized over line features we introduce the implicit function factor.




\subsubsection{Implicit Function Factor}\label{subsec:implicit-function-lines}
The standard graph-based SLAM formulation provides an elegant solution to creating a local map. However, landmarks are typically encoded as point features, which do not support encoding common shapes in the driving environment such as curved lines. These lines can be important features themselves (e.g. road lanes), or denote important boundaries between different areas (e.g. the edge between the road and a pedestrian walkway, or the edge of a building). 

In AD applications, lines are often encoded using Bézier curves, arcs, clothoids, and polylines~\cite{opendrive,Feng2022RethinkingModeling}. However, these representations do not allow combination of partially overlapping lines in an error function without interpolating, subdividing, and integrating separate measurements~\cite{MunozSanchez2022,Lu2021Graph-EmbeddedDetection}, which poses a challenge since a moving vehicle is likely observing only a part of the line. To overcome this challenge and optimize over partially observed and potentially overlapping line measurements, we introduce the \textit{implicit function factor}. 

The line is represented by an implicit equation to overcome some of the limitations of its explicit counterpart, such as support for multivalued functions, which are useful for geometries like roundabouts. Cubic polynomials are often used to describe lanes, thus the implicit function factor is defined by a bi-cubic implicit equation, and a high and low point to establish a region where this implicit equation is valid. The implicit equation is given by
\begin{equation}
\label{eq:impliciteq}
    f(x,y) =\mathbf{Q} \mathbf{c}^T = 0,
\end{equation}
where $\mathbf{c}$ and $\mathbf{Q}$ are the vector of coefficients and indeterminates, respectively, given by
\begin{equation}
\label{eq:coefficients}
    \mathbf{c} = \left[a_1, a_2, a_3, b_1, b_2, b_3, c_0, c_1, c_{12}, c_{21}\right], \text{ and}
\end{equation}
\begin{equation}
\label{eq:monomials}
    \mathbf{Q} = \left[x, x^2 , x^3, y, y^2, y^3, 1, xy, xy^2  , x^2y\right].
\end{equation}

\begin{figure}[htb]
    \centering
    \includegraphics[width=\linewidth]{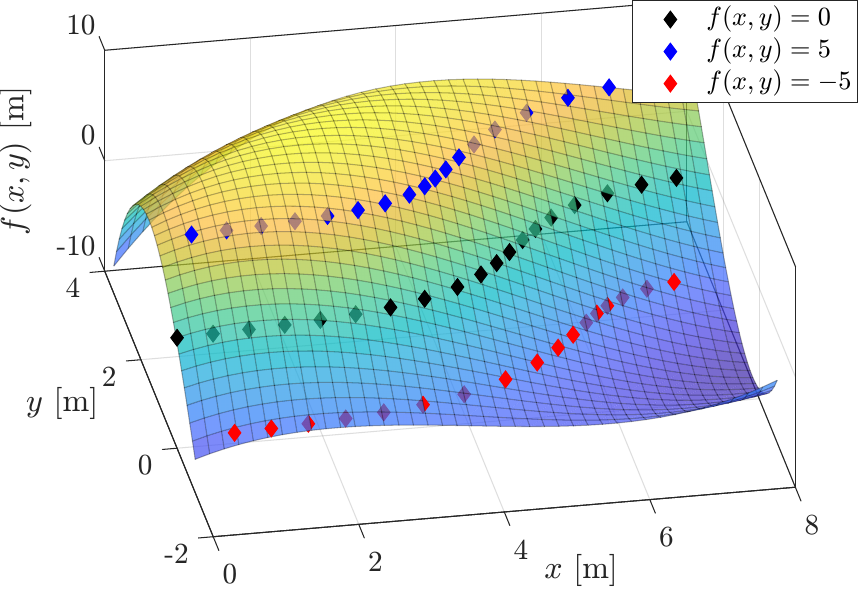}
    \caption{Example surface interpretation of implicit line function fitting using 3L, with $d_{3L}=1$ and $\|\nabla_{des}\|=5$. All line points are close to the fitted surface.}
    \label{fig:3L}
\end{figure}

Although the implicit equation defined in \eqref{eq:impliciteq} only holds for certain values of $x$ and $y$, $f(x,y)\in \mathbb{R}$ is defined for any value of $x, y \in \mathbb{R}$. The implicit function can be interpreted as a surface with height $f(x,y)$, as illustrated in Fig.~\ref{fig:3L}, which shows this surface and points of three lines at heights $f(x,y) = 0$, $f(x,y) = 5$, and $f(x,y) = -5$.

Similar to fitting a function to the data points of a line, an error function is introduced for implicit line optimization. Let the line we aim to approximate be denoted by $\Gamma$, and the length along this line by $s$, then a zeroth order distance error can be combined with a gradient error~\cite{Lei19963LPolynomials} to construct the sum of the following quadratic errors
\begin{equation}\begin{split}
\label{eq:interror-complete}
    \varepsilon = \int_{\Gamma(s)} & f(x(s),y(s))^2 \\ & + \omega (\| \nabla f(x(s),y(s))  -\nabla_{des} \|)^2 ds,
\end{split}\end{equation}
where $\| \cdot \|$ indicates the Euclidean norm, $\nabla_{des}$ is the desired gradient, and $\omega$ is a weighing variable to penalize deviations from the desired gradient. Including an objective for the gradient not only improves robustness of the fitting~\cite{Lei19963LPolynomials}, but also allows avoiding the trivial solution, where $\mathbf{c}$ is the zero vector.


Minimizing the error function given by~\eqref{eq:interror-complete} involves costly numerical evaluation of an integral, thus a more efficient function is preferred. An efficient linear approximation is the 3L method~\cite{Lei19963LPolynomials}, which uses two additional lines, $\Gamma^{+}$ and $\Gamma^{-}$, on opposing sides near the original line $\Gamma$ with level sets $\pm c_{3L}$. In Fig.~\ref{fig:3L} these are shown in blue and red, for $c_{3L} = 5$. 

The original 3L method fixes $f(\Gamma^\pm(s))$ to $\pm c_{3L}$ for the entire line. However, it is possible to obtain a more accurate fit when relaxing this constraint to an approximate value, $\Tilde{c}_{3L}$, at each point of the line~\cite{Rouhani2010RelaxingFitting}, given by:
\begin{equation}
    f(\Gamma^\pm(s)) = \pm \Tilde{c}_{3L} = \pm d_{3L}\, \|\nabla_{des}\| \approx \pm c_{3L},
\end{equation}
where $d_{3L}$ denotes the x-y planar distance between $\Gamma(s)$ and either $\Gamma^{+}(s)$ or $\Gamma^{-}(s)$. Then the new error is defined as:
\begin{equation}
    \varepsilon_{3L} = \left\|\begin{bmatrix}
        \mathbf{Q} \\  \mathbf{Q}_+ \\  \mathbf{Q}_-
    \end{bmatrix} \mathbf{c}^T - \begin{bmatrix}
        \mathbf{0} \\ \mathbf{d}_{3L} \\ \mathbf{-d}_{3L}
    \end{bmatrix} \|\nabla_{des}\| \right\| ^2, 
\end{equation}
where $\mathbf{Q}^+$ and $\mathbf{Q}^-$ denote the vector of indeterminates of $\Gamma^+$ and $\Gamma^-$, respectively. When $\mathbf{Q}$,  $\mathbf{Q}^+$ and $\mathbf{Q}^-$ are evaluated at a collection of points from $\Gamma$,  $\Gamma^+$ and $\Gamma^-$, an ordinary least squares problem results, which can be efficiently solved by inversion of the pseudoinverse~\cite{Penrose1956OnEquations}, given the linear independence of rows in $\begin{bmatrix} \mathbf{Q}$, $\mathbf{Q}^+$, $\mathbf{Q}^-\end{bmatrix}^T$.

This error function is used during optimization for line points associated with the same line feature in graph-based SLAM, which are obtained from vehicle observations for the original line $\Gamma$. Points for the adjacent lines used in the 3L method, $\Gamma^+$ and $\Gamma^-$, are obtained from the sampled points of $\Gamma$ as follows. 
For each line point $\gamma_k$, a unit vector 
perpendicular to the vector from the preceding to the subsequent points
is calculated. This unit vector is multiplied by $d_{3L}$, the Euclidean distance between $\gamma_k$ and either $\gamma^{+}_k$ or $\gamma^{-}_k$ in the x-y plane, and added and subtracted to $\gamma_k$ to calculate the adjacent points making up $\Gamma^+$ and $\Gamma^-$.

After building and optimizing the graph, as introduced in Section~\ref{subsec:slam}, and incorporating the custom implicit function factor, as detailed in Section~\ref{subsec:implicit-function-lines}, we obtain an up-to-date local map containing landmarks relative to the AV. The next step involves using the elements of this map to estimate the drivability of the space, as detailed in the following section.

\subsection{Drivability Extraction} \label{subsec:drivability}
After optimization of the graph $\mathbf{G}$, we extract the drivability of a given location, $D(x,y) \in \mathbb{R}$ in the surroundings of the AV. 
The final drivability of a location is a combination of multiple APFs derived from each node $\node\in \mathbf{G}^\mathbf{N}$, and is given by
\begin{equation}\label{eq:drivability}
    D(x,y) = \sum_{\node \in \mathbf{G}^\mathbf{N}} \mathcal{DK}_\node(L_\node(x,y)\, \cdot \, B_\node(x,y)).
\end{equation}
The drivability of entities encoded in nodes is represented with a \textit{bounding box drivability}, $B_\node(x,y)$, and an \textit{implicit line drivability}, $L_\node(x,y)$.
The geometry of objects that are typically given by their pose and bounding box (bbox)~\cite{Qian20223DSurvey} is preserved by $B_\node(x,y)$, while $L_\node(x,y)$ is used for the more complex line shapes. Finally, the drivability resulting from the aforementioned terms may be modified by applying \textit{domain knowledge} $\mathcal{DK}$, i.e. assumptions that can be made based on all available observations or prior knowledge. 
These concepts are further detailed in the next sections.

\begin{figure}[htb]
    \centering
    \includegraphics[width=\linewidth]{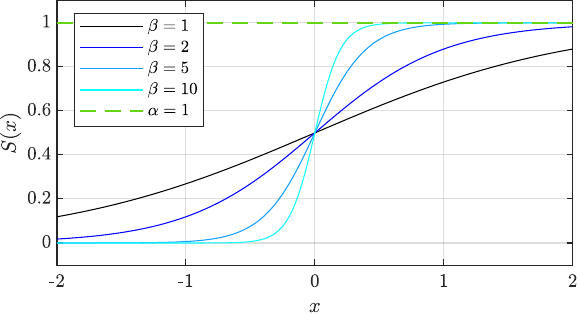}
    \caption{Sigmoid function with varying $\beta$ and $\alpha=1$.}
    \label{fig:sigmoid}
\end{figure}

To model the drivability, it is beneficial to have a flexible base function that allows separation of high and low drivability areas. To allow superposition of different drivability terms, the function should be zero over the whole domain, except for the area represented by a given node, where the value should be equal to the corresponding drivability we wish to encode. The generalized logistic function satisfies these properties~\cite{Richards1959AUse}, and setting   its left horizontal asymptote to zero and its right horizontal asymptote to the drivability amplitude $\alpha$ we obtain the standard logistic curve, or sigmoid, given by
\begin{equation}\label{eq:sigmoid}
    S(x)= \frac{\alpha}{1+ e^{-\beta \, x}}, 
\end{equation}
where $\beta$ determines the growth rate. 
An example of the resulting curves for varying $\beta$ and $\alpha=1$ can be seen in Fig.~\ref{fig:sigmoid}.
This function is symmetric around $(0,\frac{1}{2}\alpha)$, which is valuable in superposition~\cite{Ren2007AFunctions}. The resulting sigmoid satisfies the following properties:
\begin{subequations}
\label{eq:prop_sigmoid}
\begin{align}
\begin{split}
    \Lim{x\to-\infty} S(x)=0\\
    \Lim{x\to\infty} S(x)=\alpha
\end{split}\\
 & S(x)+S(-x)=\alpha \\
 & S(0)=\frac{1}{2}\alpha, \label{eq:prop_sigmoid:zero}
\end{align}
\end{subequations}
for the horizontal asymptotes, superposition principle, and the
zero-value properties, respectively.

\subsubsection{Bounding Box Drivability}
Three-dimensional objects are typically defined by their pose and their height, length and width (i.e. their bbox)~\cite{Qian20223DSurvey}. In AD applications, the shape of these objects is commonly approximated by a bi-Gaussian field~\cite{Jia2022Car-followingField,Hongyu2018AnDriving,vanderPloeg2022LongApproach,Huang2020AApproach}, which can lead to undesired behavior due to the inaccurate shape representation~\cite{vanderPloeg2022LongApproach}. For a more accurate shape representation, we combine implicit equations with the previously introduced sigmoid function, as in~\cite{Ren2007AFunctions}.

\begin{figure}[htb]
    \centering
    \includegraphics[width=\linewidth]{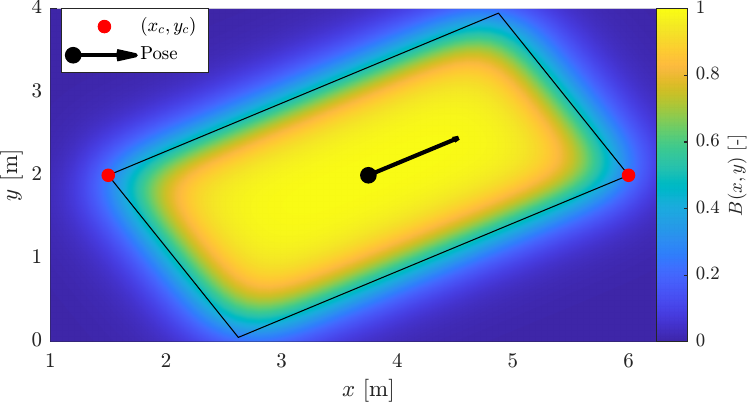}
    \caption{Bbox edges and drivability example with $\beta=5$.}
    \label{fig:BB}
\end{figure}

Suppose we wish to extract the drivability for the bbox shown in Fig.~\ref{fig:BB}, as seen from a bird's-eye view. This shape, encoded in the optimization graph node $\node$, is determined by its location $\node^{\mathbf{xy}}$, orientation $\node^{\theta}$, and size (i.e. length and width) stored in $\node^{\mathcal{S}}$. The locations of two opposite corners $(x_{c1},y_{c1})$ and $(x_{c2},y_{c2})$, as visualized by the two red points in Fig.~\ref{fig:BB}, are determined using trigonometric functions. 
The edges of the bbox then intersect with solutions to the implicit equations with implicit functions given by
\begin{multline}
\label{eq:impl_edges}
    g(x,y, i) = (x-x_c) \cos\left(\theta - \frac{i}{2} \pi\right)\\ + (y-y_c) \sin\left(\theta - \frac{i}{2} \pi\right),
\end{multline}
where $(x_c,y_c)$ is $(x_{c1},y_{c1})$ for $i=0\vee i=1$ and $(x_{c2},y_{c2})$ for $i=2\vee i=3$. Providing this function as an input to the sigmoid function results in a rise on an edge. The product of all edges results in the bbox drivability of node $\node$,  $B_\node(x,y)$, shown in Fig.~\ref{fig:BB} and given by
\begin{equation}\label{eq:multiply-edges}
    B_\node(x,y) = \prod_{i=0}^3 S(g_i(x,y)).
\end{equation}

Nodes encoding objects whose shape can be represented by a bbox alone are already well described by the bbox drivability $B_\node(x,y)$. Thus, in~\eqref{eq:drivability}, the implicit line drivability component $L_\node(x,y)$ is set to 1 for all $(x,y) \in \mathbb{R}$.  

Despite the simple example provided in Fig.~\ref{fig:BB}, it is possible to apply this technique to represent more complex shapes. Any convex shape can be created by a product of all its edges in a similar fashion as done in~\eqref{eq:multiply-edges}. Non-convex shapes can be defined by splitting them into multiple convex shapes~\cite{Ren2007AFunctions}, and adding the resulting drivabilities. 

\renewcommand{\subfigWidth}{0.195\textwidth}
\begin{figure*}[htb]
    \centering
    \begin{subfigure}[]{\subfigWidth}
      \centering
        \includegraphics[width=\textwidth]{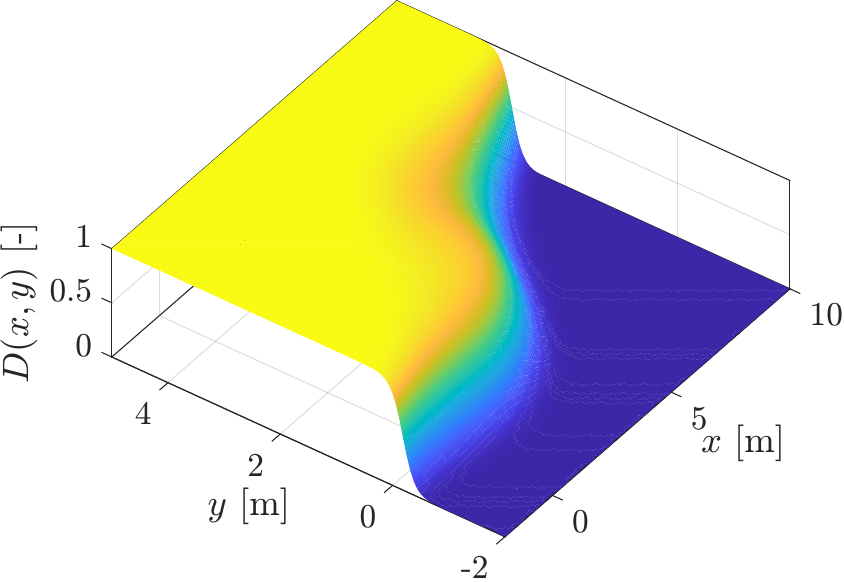}
        \caption{Implicit function sigmoid,\\$S(f(x,y))$.}
    \end{subfigure}
    \begin{subfigure}[]{\subfigWidth}
      \centering
        \includegraphics[width=\textwidth]{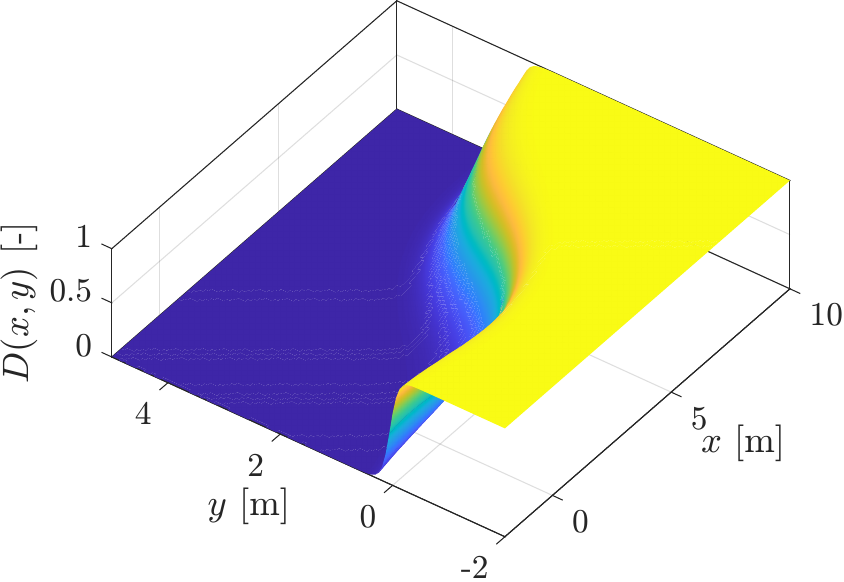}
        \caption{Negative implicit function sigmoid, $S(-f(x,y))$.}
    \end{subfigure}
    \begin{subfigure}[]{\subfigWidth}
      \centering
        \includegraphics[width=\textwidth]{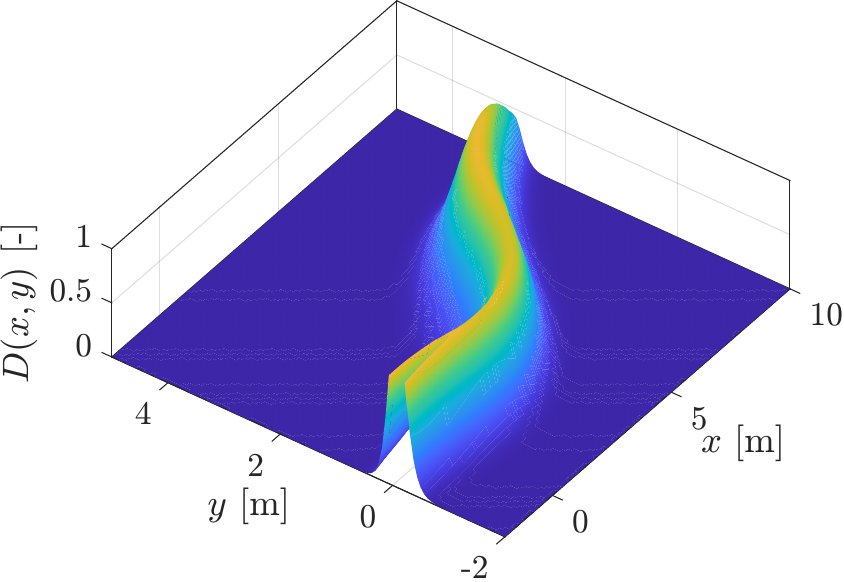}
        \caption{Product of boundary sigmoids, $S(f^l(x,y)) \cdot S(f^r(x,y))$.}
    \end{subfigure}
    \begin{subfigure}[]{\subfigWidth}
         \centering
        \includegraphics[width=\textwidth]{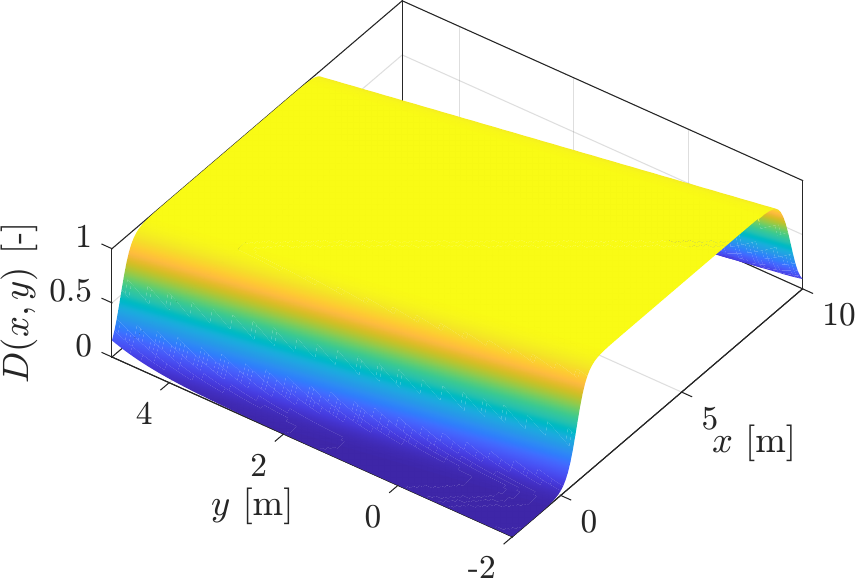}
        \caption{Bbox drivability encoding high and low bounds, $B(x,y)$.}
    \end{subfigure}
    \begin{subfigure}[]{\subfigWidth}
    \centering
    \includegraphics[width=\linewidth]{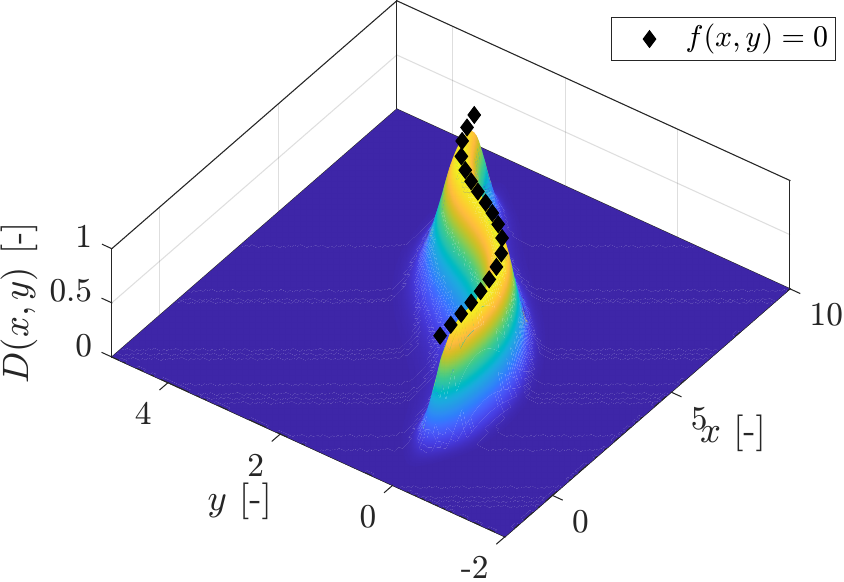}
    \caption{Drivability resulting from the product of (a)-(d).}
    \label{fig:3L_APF}
    \end{subfigure}    
    \caption{ 
    Different components of the drivability extraction for a lane line and resulting drivability.
    }
    \label{fig:DS_proces}
\end{figure*}
\subsubsection{Implicit Line Drivability}
Nodes encoding line features must also have an accurate APF representation to capture the drivability of the space surrounding them. As previously introduced in Section~\ref{subsec:implicit-function-lines}, the lines are defined by ten parameters, plus a validity region. The former is implemented via implicit functions to define the line shape, resulting in the node's line drivability $L_\node(x,y)$, while the latter is implemented with an auxiliary bbox drivability $B_\node(x,y)$ with values nearing zero in the non-valid regions, and nearing the drivability amplitude in the valid regions. 

\paragraph{Implicit function} The working principle is similar to bboxes. Since the implicit function is optimized for three level sets, it is zero where the implicit equation holds, positive on one side, and negative on the other. Recall equations~\eqref{eq:impliciteq}-\eqref{eq:monomials} and~\eqref{eq:prop_sigmoid} describing the shape of $f(x,y)$ and properties of the sigmoid $S(x)$. Their combination, $S(f(x,y))$, results in a value approaching 0 and $\alpha$ for the non-zero level sets, between which the value is $\frac{1}{2}\alpha$ for the line where the implicit equation holds, as shown by the example of Fig.~\ref{fig:DS_proces}~(a). To maintain a non-zero drivability value only for the implicit line, $S(f(x,y))$ is multiplied by a sigmoid of its negative counterpart, $S(-f(x,y))$ (Fig.~\ref{fig:DS_proces}~(b)), which results in a valid drivability along the line, leading to a shape that is similar to that depicted in Fig.~\ref{fig:DS_proces}~(c).

Note that, when multiplying $a$ sigmoids, and given \eqref{eq:prop_sigmoid:zero}, $S^a(0)$ would result in different drivability values for any $a \neq 1$, namely $\frac{1}{2^a}\alpha^a$.
Thus, to recover the original drivability amplitude, the product of $a$ sigmoids is multiplied by a factor $\phi$ given by
\begin{equation}\label{eq:phi}
    \phi = 2^a \alpha^{1-a}.
\end{equation}
Thus, when multiplying the two sigmoids $S(f(x,y))$ and $S(-f(x,y))$, to ensure $S(0) = \alpha = 1$, the product is additionally multiplied by a factor $\phi = 4$ as given by \eqref{eq:phi}.

With this operation, we ensure a rise in the value of the function at the positions represented by the desired line. However, note that the fit of $f(x,y)$ may result in solutions at undesired positions. An example of this occurrence can be seen in the top-left part of Fig.~\ref{fig:3L}, where the surface goes down and becomes 0 at positions other than $\Gamma$, which would result on high amplitude values at undesired positions after applying the sigmoid. 
To avoid non-zero values in those regions, two additional implicit functions $f^l(x,y)$ and $f^r(x,y)$ are created, reusing $\Gamma^+$ and $\Gamma-$ as boundaries, but with different level sets for the fitting. The result of combining $f^l(x,y)$ and $f^r(x,y)$ is shown in Fig.~\ref{fig:DS_proces}~(c). Combined, the line drivability of a node $\node$ is given by 
\begin{equation}
\begin{split}
    L_\node(x,y) =& \,16\, S(f_\node(x,y)) \cdot S(-f_\node(x,y))\\ & \cdot S(f_{\node}^r(x,y)) \cdot S(f_{\node}^r(x,y)).
\end{split}
\end{equation}

\paragraph{Validity region} The resulting line drivability is unconstrained along the line, as can be seen in Fig.~\ref{fig:DS_proces}~(c). To generate a drivability consistent with the partially observed line, we constrain it within the region for which we have valid line samples. To that end, from the sampled line points, a high and a low point are chosen such that the Euclidean distance between them is the largest among any two points from the line. These points are used to create a rectangular area that constrains the drivability of node $\node$ with a corresponding bounding box $B_\node(x,y)$. Since the sides of the line are already constrained by $f^l$ and $f^r$, only the edges corresponding to the high and low points are needed,
resulting in a rise as shown in Fig.~\ref{fig:DS_proces}~(d). 
 
 The resulting base drivability for a line feature after multiplication of $L_\node(x,y)$ and $B_\node(x,y)$ is shown in Fig.~\ref{fig:3L_APF}.


\subsubsection{Domain Knowledge}
To enhance drivability extraction, domain knowledge can be applied. This domain knowledge is introduced as rule-based modifications of the base drivability depending on the semantic information of the object from which drivability is extracted. Recall our definition of drivability from~\eqref{eq:drivability}, involving the $\mathcal{DK}(d_\node)$ modifier for some drivability $d_\node$ originating from node $\node$, with amplitude $d_\node^\alpha$. Domain knowledge is applied as
\begin{equation} \label{eq:dk}
    \mathcal{DK}(d_\node) = \delta_\node \, d_\node^\alpha,
\end{equation}
where $\delta_\node$ is a predefined constant depending on the node's semantic information $\node^\mathcal{S}$. With this construction the drivable space can easily encode domain knowledge rules allowing intuitive judgment of how usable the space surrounding the AV is. For instance, we can encode rules such as:

\paragraph{Avoidance of obstacles} This basic behavior can be achieved setting $\delta_{\node_i} \gg \delta_{\node_j}$ for all $\node_j \in \mathbf{N}, \node_i \neq \node_j$ if $\node_i^\mathcal{S}$ states this is a static obstacle, and $\node_j^\mathcal{S}$ states any other object.

\paragraph{Domain knowledge} If we observe vehicles driving on a certain area, we can consider that area drivable, even if that area is unknown to the AV (i.e. unmapped and currently not visible), or it is currently considered non-drivable (e.g. from an outdated map). Setting $\delta_{n} < 0$ results in an attractive drivability field allowing the AV to drive on this area. 

\paragraph{Situation-aware drivability updates} Consider the example from Fig.~\ref{fig:examples-boulder}. A conventional system would not be able to progress towards its goal, since it would not be allowed to cross the solid lane boundary. In this particular scenario, however, it would be acceptable to go around the boulder. Using the drivable space presented in this work, if an AV observes other vehicles going around the boulder, the combined behaviors of a) and b) would allow the AV to determine that, in this particular situation, planning a trajectory around the boulder is a feasible option to be considered by a behavioral motion planner.

\section{Experiments \& Results}\label{sec:experiments}
This section describes the experiments carried out to validate the performance of SDS++. 
Both simulation and real vehicle experiments were carried out, as detailed next.



\subsection{Simulation Experiments}~\label{subsec:simulation-experiments}
The performance of SDS++ is validated in CARLA~\cite{dosovitskiy2017carla}, a simulator that allows controlling challenging factors such as sensor noise and object density, while maintaining access to ground truth data. We conducted three evaluations in simulations: \textit{1)} SDS++'s ability to localize the AV under increasing noise levels, 2) the runtime impact of enlarging the optimization graph, and 3) the benefits of SDS++'s accurate object shape representation for trajectory planning.


\subsubsection{Localization accuracy \& robustness to noise}
While achieving state-of-the-art localization is not our main focus, we analyze the effects of introducing the custom implicit line factor on SDS++'s optimization process and resulting AV position estimate accuracy under variable noise levels. Noise levels in our simulations are parameterized based on data samples from the vehicle, where $0.1\sigma$ approximates the RTK-GNSS sensor accuracy (i.e. centimeter-level), and $2\sigma$ approximates GPS-IMU fusion (i.e. achievable by passenger vehicles).

We assess noise rejection capabilities for localization across different configurations: odometry only, and combinations of odometry with landmarks and/or lane lines, across five noise intensities ($0.1\sigma$ to $2\sigma$) of odometry measurements, with constant noise levels ($0.25\sigma$) for landmarks and lane lines\footnote{Note that, although the term localization typically refers to the current vehicle position, the error we compute considers the entire AV trace.}. Mean absolute error (MAE) is calculated by comparing the AV's estimated position post-optimization to the simulation's ground truth. Note that for one run, several graphs $\mathcal{G} = \{\mathbf{G}_1, \mathbf{G}_2, ..., \mathbf{G}_G  \}$ are built and optimized, one for each timestep. Thus, to calculate the localization error for the $i^{th}$ node representing an AV position in a graph, we compute 
\begin{equation}
    \mathrm{MAE}_i = \frac{1}{|\mathcal{G}'|}\sum_{\mathbf{G}\in\mathcal{G}'}\left \|\bar{\node}^{xy}_{i,\mathbf{G}}\, - \, \node^{xy}_{i,\mathbf{G}}\right \|,
\end{equation}
where $\mathcal{G'}$ denotes the optimization graphs that contain node $\node_i$, that is, $\mathcal{G'}=\{\mathbf{G} \mid \node_i \in \mathbf{G}^\mathbf{N}  \}$; $\bar{\node}^{xy}_{i,\mathbf{G}}$ denotes the ground truth position of node $\node_i$ from graph $\mathbf{G}$, and $\node^{xy}_{i,\mathbf{G}}$ denotes the optimized position of node $\node_i$ from graph $\mathbf{G}$. 

\begin{figure}[htb]
    \centering
    \includegraphics[width=\linewidth]{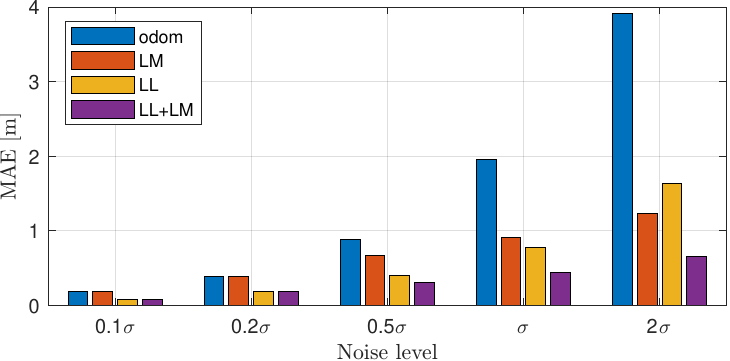}
    \caption{Localization MAE for five noise levels, where landmark (LM), lane lines (LL), or both complement the odometry measurements of the graph.}
    \label{fig:errorbar}
\end{figure}

The average localization MAE per noise level is detailed in Fig.~\ref{fig:errorbar}, displaying four groups: optimizing using odometry alone, and odometry combined with LM nodes, LL nodes, or both. Odometry-only graphs act as a baseline, showing unoptimized MAE levels. Performance improvement with LMs is only notable when odometry noise surpasses landmark noise, while LLs benefit localization even at lower noise levels like $0.1\sigma$ and $0.2\sigma$.
\renewcommand{\subfigWidth}{0.329\textwidth}
\begin{figure*}[htb]
    \centering
    \begin{subfigure}[t]{\subfigWidth}
      \centering
        \includegraphics[width=\textwidth]{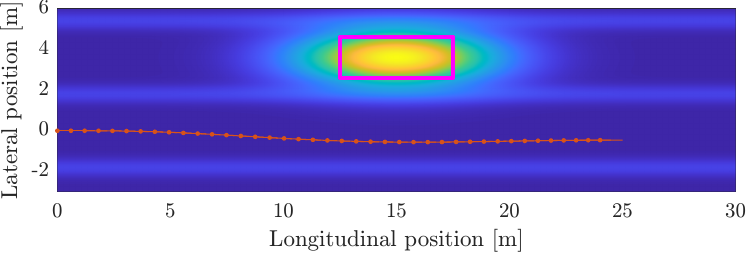}
        \caption{Gaussian APF for a vehicle in the adjacent lane}\label{fig:improvedBB_gaus_adj_large}
    \end{subfigure}
    \begin{subfigure}[t]{\subfigWidth}
      \centering
        \includegraphics[width=\textwidth]{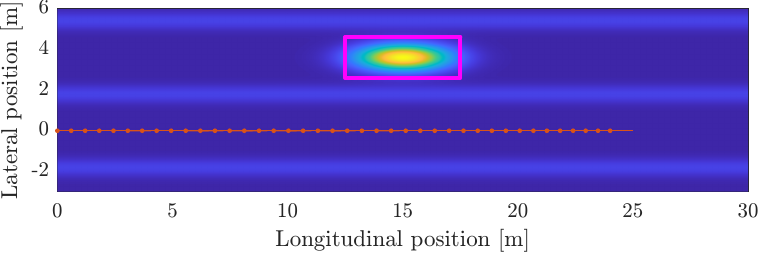}
        \caption{Gaussian APF with decreased spread for a vehicle in the adjacent lane}\label{fig:improvedBB_gaus_adj_small}
    \end{subfigure}
    \begin{subfigure}[t]{\subfigWidth}
      \centering
        \includegraphics[width=\textwidth]{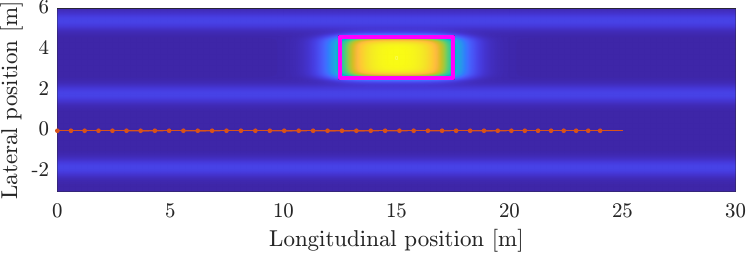}
        \caption{SDS++ APF for a vehicle in the adjacent lane}\label{fig:improvedBB_sds_adj}
    \end{subfigure}
    \begin{subfigure}[t]{\subfigWidth}
      \centering
        \includegraphics[width=\textwidth]{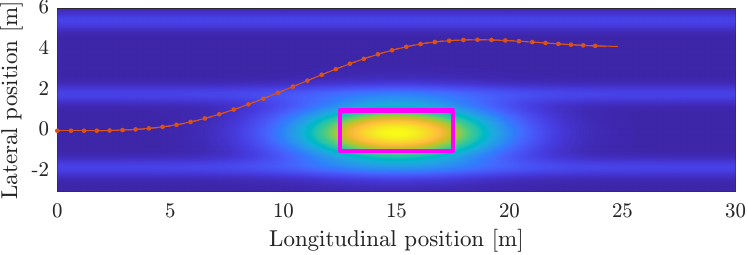}
        \caption{Gaussian APF for a vehicle in the AV lane}\label{fig:improvedBB_gaus_ego_large}
    \end{subfigure}
    \begin{subfigure}[t]{\subfigWidth}
      \centering
        \includegraphics[width=\textwidth]{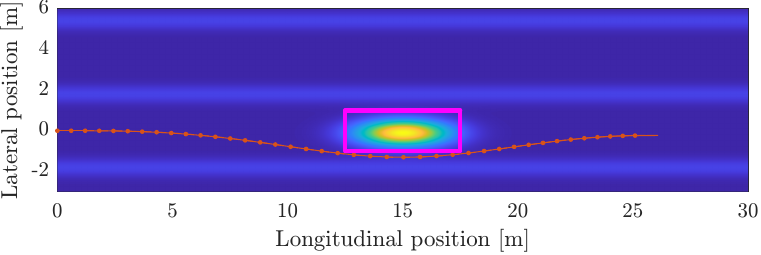}
        \caption{Gaussian APF with decreased spread for a vehicle in the AV lane}\label{fig:improvedBB_gaus_ego_small}
    \end{subfigure}
    \begin{subfigure}[t]{\subfigWidth}
         \centering
        \includegraphics[width=\textwidth]{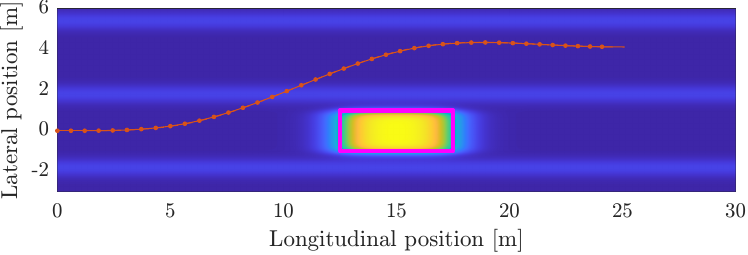}
        \caption{SDS++ APF for a vehicle in the AV lane}\label{fig:improvedBB_sds_ego}
    \end{subfigure}
    \caption{
    Comparison of the planned trajectory using Gaussian and SDS++'s object shape representation in a two-way road scenario where a vehicle blocks either the AV lane (bottom) or the adjacent lane (top). 
    }
    \label{fig:improvedBB}
\end{figure*}

Despite identical noise settings for LLs and LMs, LLs consistently show better MAE at noise levels below $0.5\sigma$, leveraging multiple line points and implicit orientation data. However, as noise increases beyond $\sigma$, LLs' effectiveness declines due to the limitations of the implicit function setup, potentially leading to local minima.

Furthermore, high noise levels or poor initial conditions are known to significantly impair the performance of GTSAM relative to other optimization back-ends~\cite{Juric2021ASLAM}, resulting in higher MAEs at elevated noise levels. Although achieving state of the art localization accuracy is not our primary goal, our results confirm that integrating the implicit function factor for line features enhances the optimization process across all noise levels, as shown by the lowest MAEs achieved when using all measurements.

\subsubsection{Accurate Object Shape Representation}
To illustrate the benefits of the more accurate object representation provided by SDS++, we compare it to the commonly used bi-Gaussian representation in a simulated environment with a one-way, two-lane straight road, inspired by~\cite{vanderPloeg2022LongApproach}. Two variations are tested. In the first, a vehicle is blocking the AV lane, and in the second, the vehicle is in the adjacent lane. 

When comparing the representation given by the bi-Gaussian to that of SDS++, a trajectory planner is used to analyze the trajectories the AV would plan given both representations. The specific planner used is further detailed in the next section. 

Fig.~\ref{fig:improvedBB} illustrates the benefits of improved object representation accuracy in a two-lane road, where a vehicle is stationary in the AV's lane (right), or the adjacent lane (left). The trajectory that planner computes is shown in orange, and the vehicle's bbox is indicated by the magenta edges. In Figures~\ref{fig:improvedBB_gaus_adj_large}, ~\ref{fig:improvedBB_gaus_adj_small}, ~\ref{fig:improvedBB_gaus_ego_large}, and \ref{fig:improvedBB_gaus_ego_small} a bi-Gaussian is used to represent the vehicle, where the standard deviation is larger for \ref{fig:improvedBB_gaus_adj_large} and \ref{fig:improvedBB_gaus_ego_large} than for \ref{fig:improvedBB_gaus_adj_small} and \ref{fig:improvedBB_gaus_ego_small}. Figures~\ref{fig:improvedBB_sds_adj} and ~\ref{fig:improvedBB_sds_ego} use SDS++'s drivability representation. 

When the vehicle is on the adjacent lane and represented with the commonly used bi-Gaussian and larger standard deviation values, although the planner computes a safe trajectory in the AV's lane, there is unnecessary swerving (Fig.~\ref{fig:improvedBB_gaus_adj_large}). This unnecessary swerving can be avoided lowering the standard deviation, as shown in Fig.~\ref{fig:improvedBB_gaus_adj_small}, however, this representation can lead to potentially unsafe trajectories that are too close to the vehicle (Fig.~\ref{fig:improvedBB_gaus_ego_small}). The proposed bbox drivability of SDS++ allows avoiding this kind of undesired swerving behavior (Fig.~\ref{fig:improvedBB_sds_adj}), as well as ensuring the object's shape is accurately represented to avoid getting too close (Fig.~\ref{fig:improvedBB_sds_ego}).
%


\subsubsection{Runtime}\label{subsubsec:runtime} 
The running time of SDS++ is influenced by several factors such as hardware specifics, efficiency of the implementation, and noise levels in the data. The former two are fixed, and to minimize the variability of the latter, we maintain a constant noise level of $0.2\sigma$ across all graph nodes and evaluate the execution time across three phases: constructing the graph, optimizing the graph, and constructing the APFs for drivability encoding. These timing experiments were performed on a modest laptop equipped with an Intel Core i7-7700HQ CPU.

\begin{figure}[htb]
    \centering
    \includegraphics[width=1\linewidth]{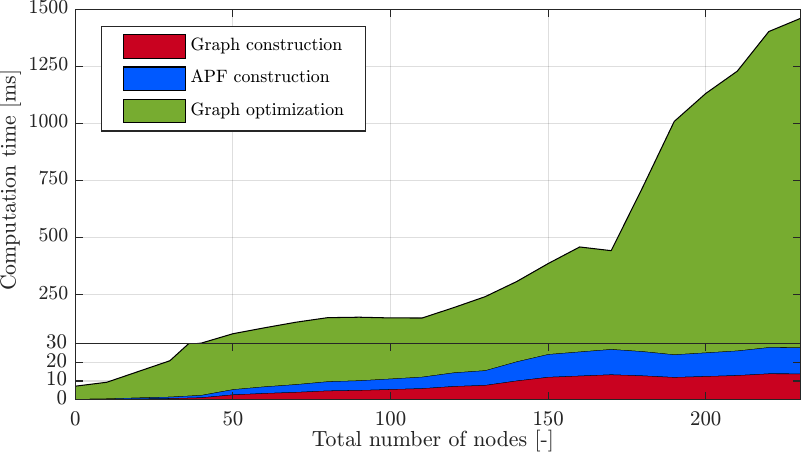}
    \caption{Runtime of SDS++.}
    \label{fig:runtime}
\end{figure}

Fig.~\ref{fig:runtime} illustrates the running time of SDS++ relative to the number of nodes in the graph, categorized by activity type. As expected, there is a positive correlation between the number of graph elements and the running time.

In the current unoptimized implementation, SDS++ maintains a running time below 250ms for up to 125 nodes, but this increases rapidly beyond that. In practical deployment, a time-based sliding window limits the number of objects considered during optimization. Most graph elements originate from efficient AV and LM observations, with fewer from the computationally intensive line detections, which have a greater impact on running time due to their complex error functions.

In addition to the simulations, real-world applicability was tested with vehicle data, where SDS++ operated at 4Hz on a modest laptop, demonstrating its efficiency and suitability for real-world applications.

\subsection{Vehicle Experiments}~\label{subsec:vehicle-experiments}
\begin{figure}[htb]
    \centering
    \includegraphics[width=\linewidth]{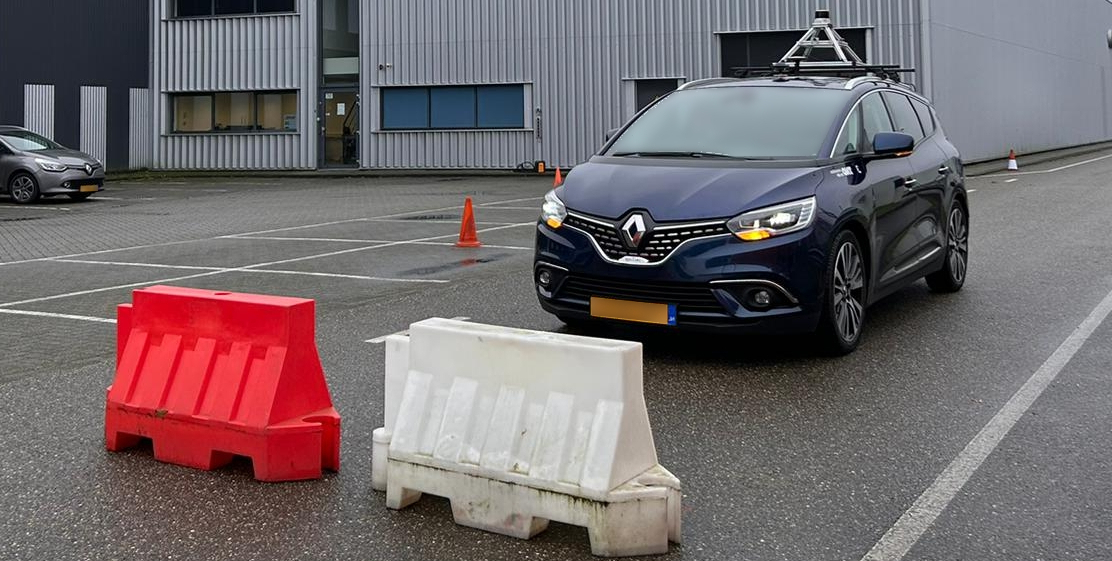}
    \caption{AD-equipped test vehicle encountering an obstruction.}
    \label{fig:carlab}
\end{figure}

To verify SDS++'s performance, we conducted experiments using the AD-equipped vehicle shown in Fig.~\ref{fig:carlab}, fitted with various sensors detailed below. This section outlines the scenarios and planner used for assessment, demonstrating how SDS++ enables adaptive trajectory planning. Note that, although the experiments utilized vehicle data, the AV was driven manually to gather data, which was then used as input to SDS++ and the planner to analyze the generated trajectories.


\subsubsection{The vehicle}\label{subsec:vehicle}
The AV is equipped with three sensors for our experiments: an \textit{Ouster OS2} LiDAR on the roof, a \textit{Mobileye} camera \cite{Real-TimeAU}, and an RTK-GNSS from \textit{OxTS}~\cite{OxTS1998}. The LiDAR data is processed by Pointpillars~\cite{Lang2019PointPillars:Clouds}, a pretrained neural network for object detection, and the Mobileye camera is mainly used for lane detection. All tracking of detected objects is performed using an extended Kalman filter \cite{Thrun2002ProbabilisticRobotics}.

\subsubsection{Scenarios}\label{subsec:scenarios}
\begin{figure}[htb]
    \centering
    \includegraphics[width=\linewidth]{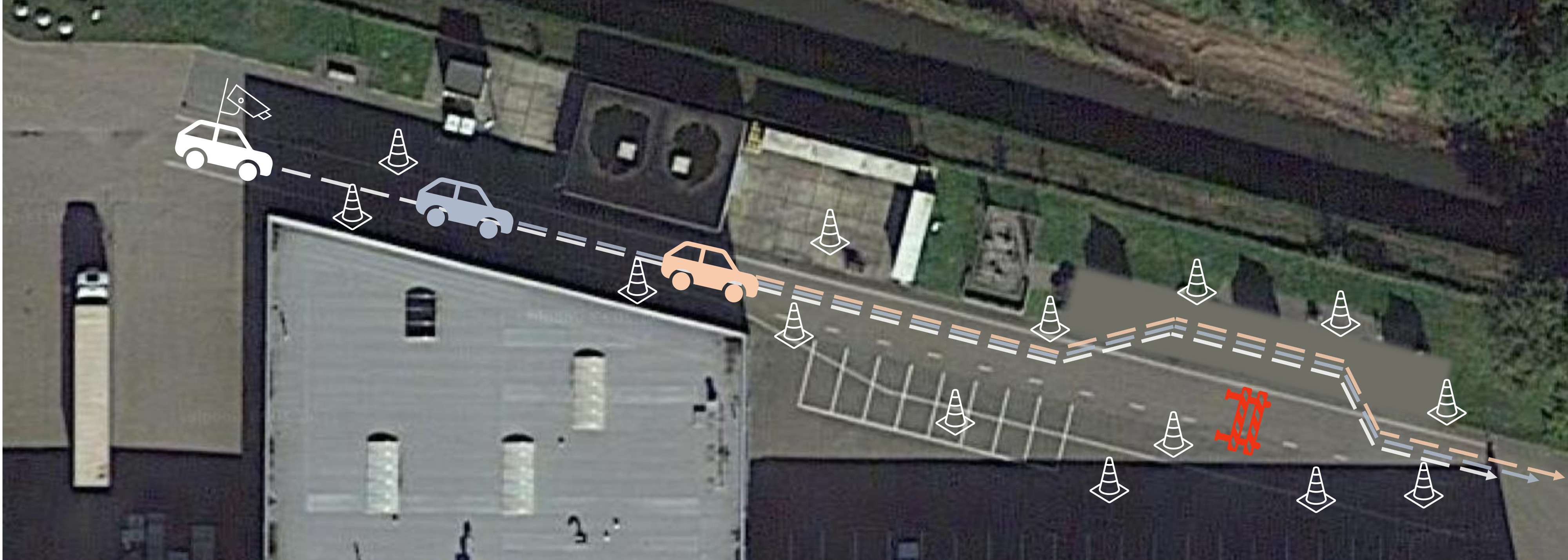}
    \caption{Birdseye view of the experiment location and instructed driving routes for the swerving scenario.}
    \label{fig:scenario_bev}
\end{figure}

Data collection for the vehicle experiments is carried out with following scenarios in a straight 2-lane road:
\begin{itemize}[leftmargin=15pt]
    \item \textit{Base}: The AV drives without other vehicles or obstructions.
    \item \textit{Follow}: The AV follows a vehicle without any obstructions.
    \item \textit{Barrier}: The AV drives towards an obstruction (i.e. a barrier) and stops in front of it. 
    \item \textit{Stop}:  The AV follows a vehicle that stops due to an obstruction, and the AV stops behind it. 
    \item \textit{Parked}: The AV drives without obstructions, but a vehicle is parked in the adjacent lane.
    \item \textit{Parking}: The AV follows a vehicle that swerves around an obstruction and parks next to it.
    \item \textit{Swerving}: The AV follows two vehicles; all three swerve around an obstruction (see Fig.~\ref{fig:scenario_bev}). 
\end{itemize}
Since a map of the testing area is not available, cones are placed to serve as static landmarks in all scenarios. Fig.~\ref{fig:scenario_bev} provides a bird's eye view of the testing site and the driving routes for the swerving scenario.

\subsubsection{Planner}\label{subsec:planner}
To verify the trajectory planned by the AV using the drivable space estimated by SDS++, we implemented a model predictive control (MPC) trajectory planner (MPTP)~\cite{Jia2022Car-followingField, Bhatt2023MPC-PF:Fields, vanderPloeg2022LongApproach}. Recorded vehicle data was replayed and inputted into SDS++ to estimate the drivable space, which MPTP then used to plan a trajectory.

The implemented MPTP is heavily inspired by~\cite{vanderPloeg2022LongApproach}. The authors of~\cite{vanderPloeg2022LongApproach} optimize over a cost function that includes standard terms in MPC optimization, like penalization of control errors and inputs, as well as deviations from a desired vehicle state. Additionally, they include two additional terms in the form of a risk field for other road users and the road infrastructure. These terms are analogous to our drivable space, and are replaced by the drivability given by SDS++.

Table~\ref{tab:MPTP_param} details the specific parameterization of SDS++ and MPTP used during the experiments.




\begin{table}[htb]
\centering
\caption{SDS++ and MPTP parameters used for evaluation}
\begin{adjustbox}{max width=\linewidth}
\begin{tabular}{ll}
\toprule
\textbf{Parameter}           & \textbf{Value} \\ \hline
 Drivability modifier of static objects ($\delta_\node$ in \eqref{eq:dk})  & $1\times10^5$\\
 Drivability modifier of solid lane line ($\delta_\node$ in \eqref{eq:dk}) &  $1\times10^3$ \\
Gradient of static objects ($\beta$ in \eqref{eq:sigmoid})    & 10   \\
 Gradient of solid lane line ($\beta$ in \eqref{eq:sigmoid})    & 5    \\
 Acceleration input limits (min, max) & (-5, 1)  m s$^{-2}$\\
 Steering velocity input limits (min, max) &  (-0.2, 0.2) rad \\
 Steering angle limits (min, max)  &  (-0.5, 0.5) rad\\
 Velocity limits (min, max)   & (0, 10]) m s$^{-1}$ \\
 Horizon length, $N_h$ & 40 \\ 
 Sample time, $t_s$ & 0.2 s\\ 
 Goal ahead**, $x_{goal}$ & $x_0+15+0.5\, v_0\, t_s\, N_{h}$ m\\
 Reference state* & [0, 0, 0, 3, 0]\\ 
 Terminal state* & [$x_{goal}$, 0, 0, 0, 0]\\  
 Input weights   & [4, 10]\\ 
 Reference state* weights & [0, 0, 0, 0.1, 0]\\ 
 Terminal state* (N) weights & [3, 2, 10, 10, 0]\\ 
 
 Car length  & 4m \\ \bottomrule
 \multicolumn{2}{p{\linewidth}}{*\scriptsize Vehicle is modeled by a bicycle model. States are given by longitudinal and lateral position, heading angle, velocity, and steering angle.}  \\
 \multicolumn{2}{p{\linewidth}}{**\scriptsize $x_0$: initial AV logitudinal position, $v_0$: initial AV speed.}  \\
\end{tabular}
\end{adjustbox}

\label{tab:MPTP_param}
\end{table}

\subsubsection{Adaptive Trajectory Planning with SDS++}~\label{subsec:results-planning}
This section highlights the adaptive capabilities of SDS++ combined with MPTP in selected real vehicle data scenarios. The \textit{follow} and \textit{parked} scenarios (Fig.\ref{fig:MPTP_follow_parking}) demonstrate the AV's behavior under normal conditions, while the \textit{swerving} scenario (Fig.\ref{fig:MPTP_swerving}) showcases its ability to adapt and navigate around obstacles that would typically halt the AV using traditional mapping methods.

In the figures, circular high-amplitude areas represent cones used as static landmarks, and rectangular high-amplitude areas indicate other static objects like static vehicles and barriers (Fig.~\ref{fig:MPTP_parking}). The planned trajectory by MPTP is shown in orange, and darker blue areas indicate regions where a vehicle has been observed driving, thus confirming their drivability. 
Note that coordinate system is relative to the first AV position in the graph\footnote{Recall that a sliding window is used to discard old nodes, in this case older than 10 seconds.}, and hence in Fig.~\ref{fig:MPTP_swerving} the object positions seem to differ between the subfigures.

Fig.\ref{fig:MPTP_follow} shows MPTP's plan at 15.4s in the \textit{follow} scenario. The darker blue area indicates a more drivable region after observing a vehicle passing through, though this vehicle is not visible since it is moving and not detected as a static obstacle. In the \textit{parking} scenario, the AV follows a vehicle that swerves around an obstacle and stops on the road shoulder. Fig.\ref{fig:MPTP_parking} shows this at 23.7s, where the AV plans to stop in front of the obstacle, unable to find a suitable trajectory consistent with the current drivable space. Note that MPTP does not consider the AV's geometry for simplicity, and it plans the trajectory until encountering the obstruction.
\renewcommand{\subfigWidth}{0.95\linewidth}
\begin{figure}[tb]
    \centering
    \begin{subfigure}[]{\subfigWidth}
      \centering
        \includegraphics[width=\textwidth]{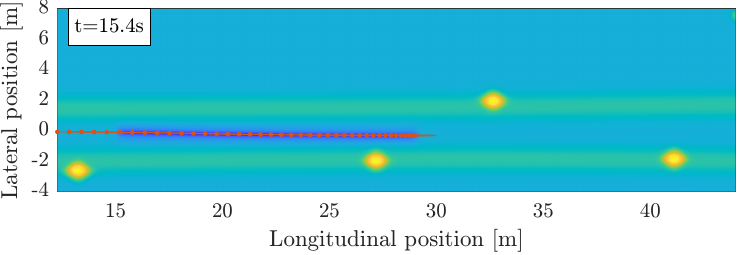}
        \caption{Scenario \textit{follow}.}
    \label{fig:MPTP_follow}
    \end{subfigure}
    \begin{subfigure}[]{\subfigWidth}
      \centering
        \includegraphics[width=\textwidth]{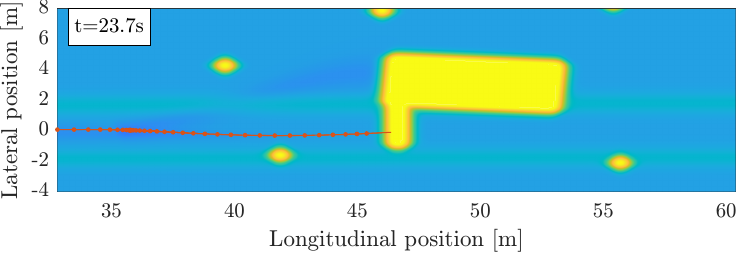}
        \caption{Scenario \textit{parking}. The parked car is not followed, and the MPTP stops in lane.}
    \label{fig:MPTP_parking}
    \end{subfigure}
    \caption{MPTP's planned trajectory during two scenarios showing that the vehicle stays in the lane.}
    \label{fig:MPTP_follow_parking}
\end{figure}

The \textit{swerving} scenario, depicted in Fig.\ref{fig:MPTP_swerving}, introduces a variation where preceding vehicles swerve around an obstacle and continue driving. Initially (Fig.\ref{fig:MPTP_swerving_1}), one vehicle has high drivability due to its stationary status, causing MPTP to plan a stop. As it moves, its bbox is no longer considered an obstacle, allowing the AV to proceed. At 21.7s (Fig.\ref{fig:MPTP_swerving_2}), the AV observes the obstacle and the vehicles in front, recognizing the drivable area and planning a trajectory. At 28.3s (Fig.\ref{fig:MPTP_swerving_3}), the AV leverages the behavior of other vehicles to cross a solid lane boundary and bypass the obstacle. Finally, at 35s (Fig.~\ref{fig:MPTP_swerving_4}), MPTP plans a trajectory to return to the original lane.


\renewcommand{\subfigWidth}{0.95\linewidth}
\begin{figure}[htb]
    \centering
    \begin{subfigure}[]{\subfigWidth}
      \centering
        \includegraphics[width=\textwidth]{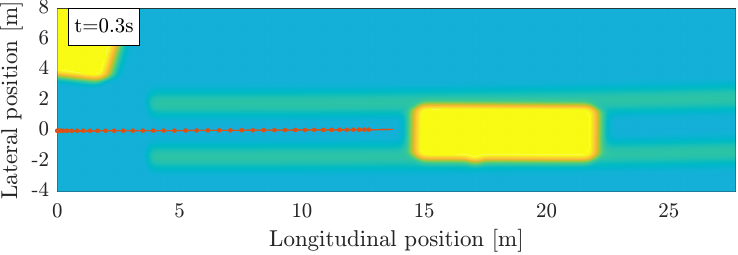}
        \caption{}
        \label{fig:MPTP_swerving_1}
    \end{subfigure}
    \begin{subfigure}[]{\subfigWidth}
      \centering
        \includegraphics[width=\textwidth]{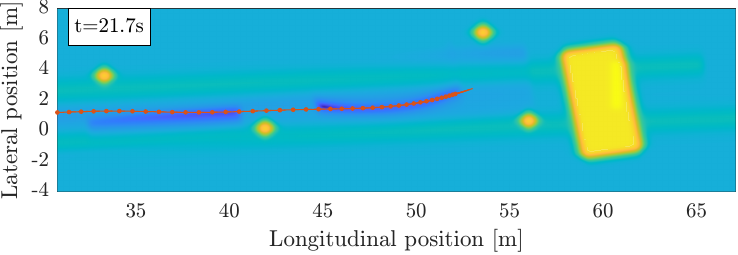}
        \caption{}\label{fig:MPTP_swerving_2}
    \end{subfigure}
    \begin{subfigure}[]{\subfigWidth}
      \centering
        \includegraphics[width=\textwidth]{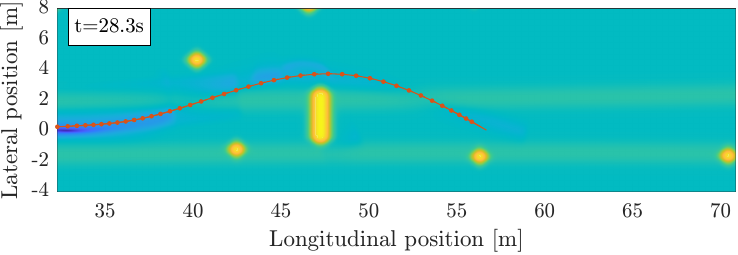}
        \caption{}\label{fig:MPTP_swerving_3}
    \end{subfigure}
    \begin{subfigure}[]{\subfigWidth}
         \centering
        \includegraphics[width=\textwidth]{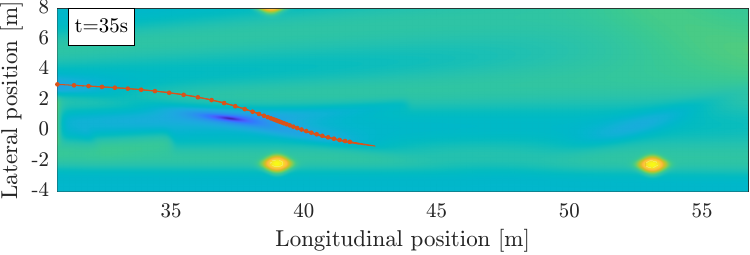}
        \caption{}\label{fig:MPTP_swerving_4}
    \end{subfigure}
    \caption{AV encountering an obstruction and planning to swerve using domain knowledge for scenario \textit{swerving}.}
    \label{fig:MPTP_swerving}
\end{figure}

\section{Conclusion}\label{sec:conclusion}
Autonomous Vehicles (AVs) require accurate and up-to-date environmental representations to navigate safely. Traditional methods relying on detailed, static maps often struggle with dynamically changing conditions or outdated data, emphasizing the need for real-time solutions that can integrate diverse data sources and adapt to current driving situations.


This paper presents SDS++, an enhancement of the SDS (situation-aware drivable space). SDS++ addresses the limitations of SDS while preserving its advantages by transitioning from a graph-based representation to artificial potential fields (APFs). This shift offers a more standardized approach to modeling drivable space, which is essential for dynamic, real-time navigation.

SDS++ also incorporates an error function for direct optimization of line features in graph-based SLAM, enhancing the robustness and accuracy of the environmental model. Additionally, SDS++ represents objects as APFs using implicit and sigmoid functions, enabling more accurate modeling of complex geometries compared to the commonly used bi-Gaussian representation. This improvement is crucial for precise trajectory planning and adaptability in unpredictable real-world scenarios.

In testing, SDS++ has been integrated with a model predictive control-based planner, showing significant improvements in trajectory planning by adjusting to current driving conditions and increasing robustness against localization noise. These results, validated through simulations and real-world data, confirm the effectiveness and reliability of SDS++.

Looking forward, future work will focus on developing data-driven methods to automate the parameterization of SDS++ and further refining its integration with planning systems across a variety of challenging real-life scenarios.

\printbibliography

\newpage

\section{Biography}
 

\newcommand{\negSpace}{-25pt}

\vspace{\negSpace}
\begin{IEEEbiography}[{\includegraphics[width=1in,height=1.25in,clip,keepaspectratio]{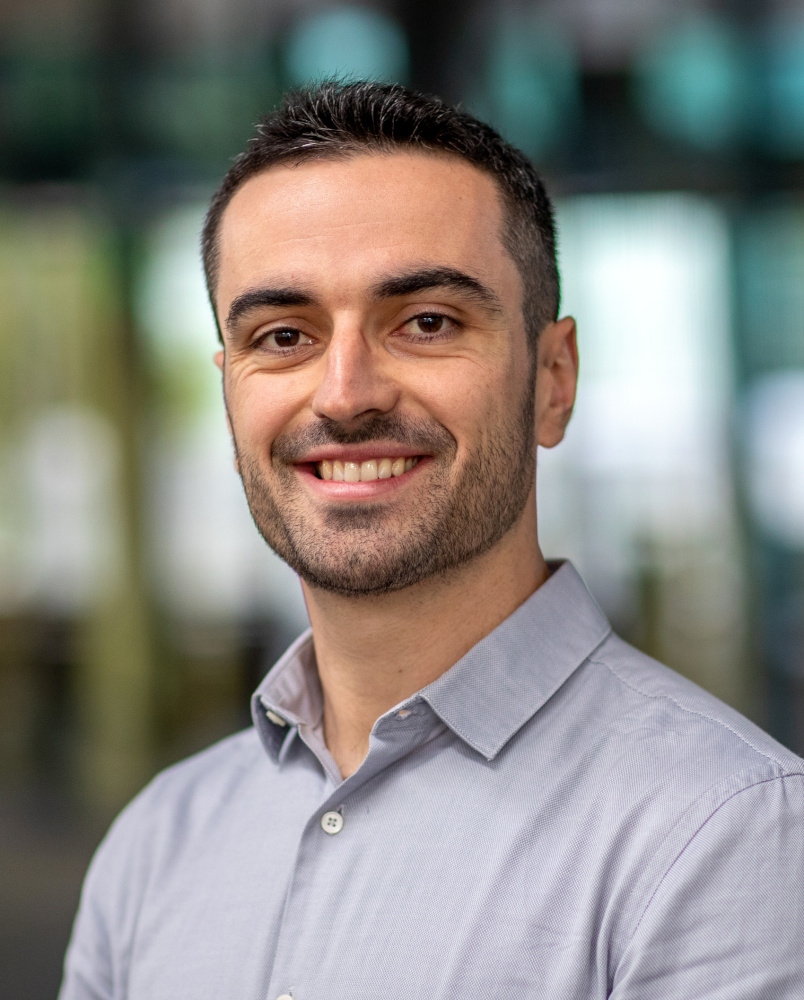}}]{Manuel Muñoz Sánchez}
received the B.Sc. and M.Sc. degrees in Computer Science and Engineering from the Eindhoven University of Technology, Eindhoven, The Netherlands, in 2017 and 2020, respectively. He is currently pursuing a Ph.D. degree in the Department of Mechanical Engineering at the Eindhoven University of Technology. His main research interest is in the applications of deep learning methods for automated driving, including thorough evaluations of trajectory prediction models and their integration with the rest of the autonomy stack.
\end{IEEEbiography}

\vspace{\negSpace}
\begin{IEEEbiography}[{\includegraphics[width=1in,height=1.25in,clip,keepaspectratio]{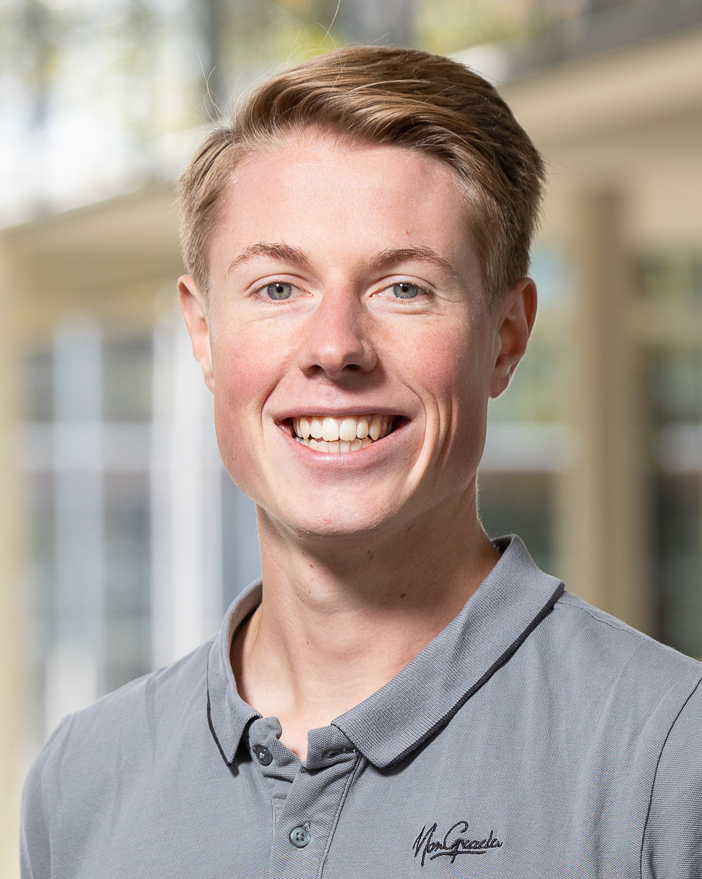}}]{Gijs Trots} received his B.Sc. and M.Sc. degrees in Mechanical Engineering, with specialization in Control Systems Technology from Eindhoven University of Technology, Eindhoven, in 2021 and 2024, respectively. During the M.Sc. studies he has done research at the Netherlands Organization for Applied Scientific Research (TNO), Helmond, the Netherlands.
\end{IEEEbiography}

\vspace{\negSpace}
\begin{IEEEbiography}[{\includegraphics[width=1in,height=1.25in,clip,keepaspectratio]{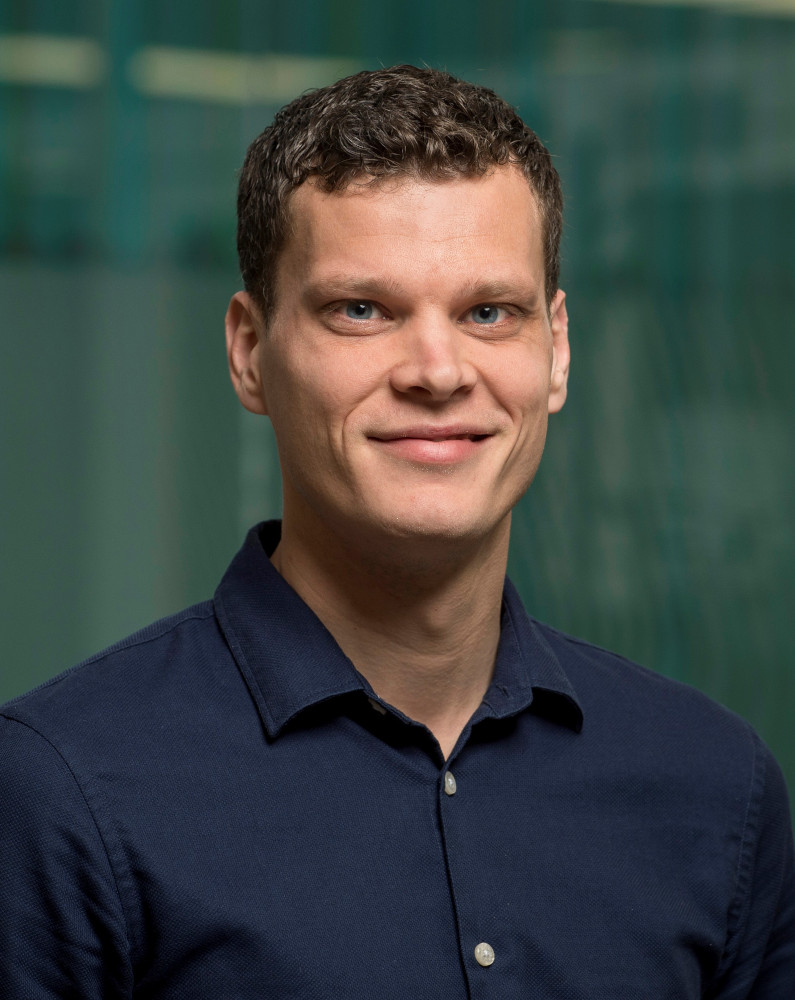}}]{Robin Smit}
received the B.Sc. and M.Sc. degrees in Mechanical Engineering and System \& Control from the Eindhoven University of Technology, Eindhoven, The Netherlands, in 2013 and 2016, respectively. Since 2019, he has been a research scientist with the Netherlands Organization for Applied Scientific Research (TNO), Helmond, The Netherlands. His work focuses on perception technologies of Automated Vehicles, including localization, ego state estimation, object tracking and prediction and road modelling. 
\end{IEEEbiography}

\vspace{\negSpace}
\begin{IEEEbiography}[{\includegraphics[width=1in,height=1.25in,clip,keepaspectratio]{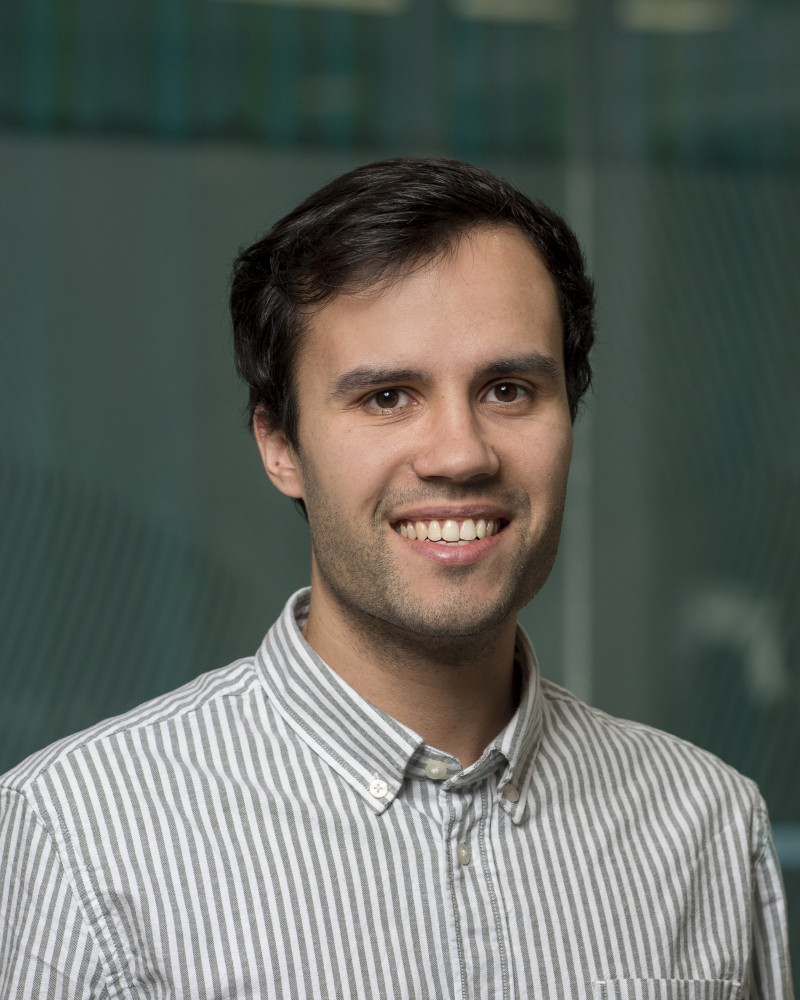}}]{Pedro Vieira Oliveira}received the B.Sc. in Mechanical Engineering at the University of Coimbra and M.Sc. degree in Automotive Technology at the Eindhoven University of Technology, in 2019 and 2022, respectively. Since 2022, he has been a Research Scientist with the Integrated Vehicle Safety Department, Netherlands Organisation for Applied Scientific Research (TNO), Helmond, The Netherlands.
\end{IEEEbiography}

\vfill
\newpage
\renewcommand{\negSpace}{-10pt}

\vspace{\negSpace}
\begin{IEEEbiography}[{\includegraphics[width=1in,height=1.25in,clip,keepaspectratio]{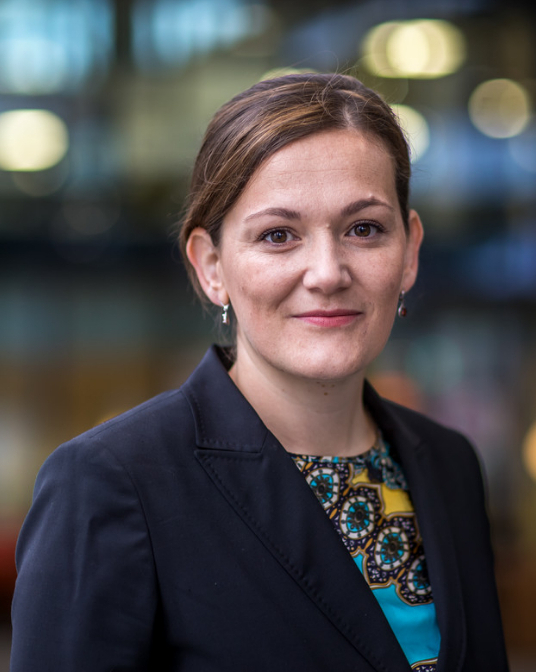}}]{Emilia Silvas}
received the B.Sc. degree in automatic control and computer science from Politehnica University of Bucharest, Romania, in 2009 and the M.Sc. degree in systems and control and the Ph.D. degree from Eindhoven University of Technology, Eindhoven, The Netherlands, in 2011 and 2015, respectively. Since 2016, she has been a Research Scientist with the Netherlands Organization for Applied Scientific Research (TNO), Helmond, The Netherlands, focusing on cooperative vehicle systems and mobile robots, and leading the Smart Vehicles programs Cluster. Her research interests include advanced control, system identification and modeling, machine learning techniques, and optimal system design.  In 2020, she became Chair of the Mobility, Transport and Logistics working group at the Dutch AI Coalition.
\end{IEEEbiography}

\vspace{\negSpace}
\begin{IEEEbiography}[{\includegraphics[width=1in,height=1.25in,clip,keepaspectratio]{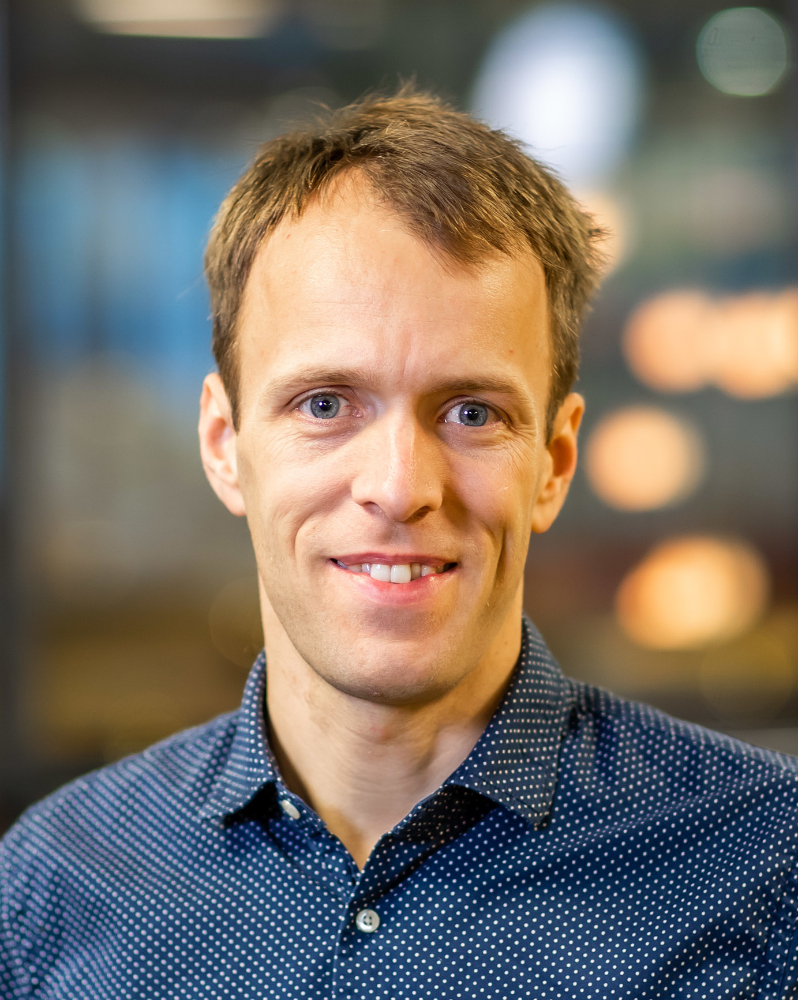}}]{Jos Elfring}
is an assistant professor at Eindhoven University of Technology. He received his MSc in Mechanical Engineering from TU/e in 2009 and his PhD in 2014. He successfully finished the Dutch Institute of Systems and Control course program in 2011, was a member of the Tech United RoboCup@Home team for several years and won the best paper award at the 2013 RoboCup Symposium. His research interests include mapping, localization, motion prediction, and data association.
\end{IEEEbiography}

\vspace{\negSpace}
\begin{IEEEbiography}[{\includegraphics[width=1in,height=1.25in,clip,keepaspectratio]{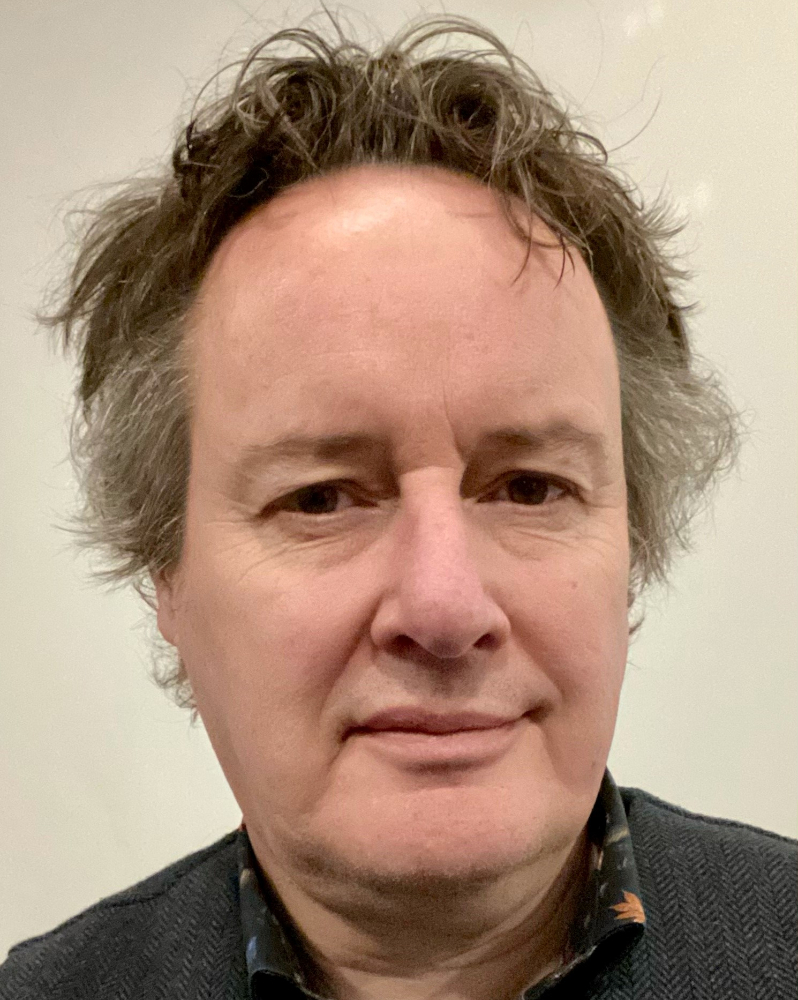}}]{René van de Molengraft}
is an Associate Professor at the Faculty of Mechanical Engineering at Eindhoven University of Technology. In 2005, he founded the Tech United Robocup team (world champion MSL in 2012, 2014, 2016, 2018, 2019 and world champion @Home 2019). He was coordinator of the European FP7 project named RoboEarth. His main research interest is world modelling and control for autonomous robots. 
\end{IEEEbiography}




\vfill

\end{document}